\documentclass[11pt]{article}

\usepackage[final]{acl}

\usepackage{times}
\usepackage{latexsym}
\usepackage{amsmath}
\usepackage{booktabs}
\usepackage{multirow}
\usepackage{amssymb}
\usepackage{pifont}     % 提供叮当字体符号
\usepackage{algorithm}
\usepackage{makecell}
\usepackage{algpseudocode}
\usepackage[table]{xcolor}
\definecolor{tfcolor}{gray}{0.93}
\newcommand{\tfc}{\cellcolor{tfcolor}}
\newcommand\blfootnote[1]{%
  \begingroup
  \renewcommand\thefootnote{}\footnote{#1}%
  \addtocounter{footnote}{-1}%
  \endgroup
}
\usepackage[T1]{fontenc}
\usepackage[utf8]{inputenc}

\usepackage{microtype}

\usepackage{inconsolata}

\usepackage{graphicx}

\title{Routing Before Looking: Query-Adaptive Evidence Acquisition for Long-form Video Understanding}

\author{
  \textbf{Tianyue Wang\textsuperscript{1,2,3}} \quad
  \textbf{Xuying Wu\textsuperscript{4}} \quad
  \textbf{Yuxiang Ma\textsuperscript{5}} \quad
  \textbf{Ruiming Liang\textsuperscript{2}} \quad
  \textbf{Jiaxuan Kang\textsuperscript{3}} \\
  \textbf{Yanchao Hao\textsuperscript{3}$^\dagger$} \quad
  \textbf{Zheng Wei\textsuperscript{3}} \quad
  \textbf{Leigang Qu\textsuperscript{6}$^\ast$} \quad
  \textbf{Haiyun Guo\textsuperscript{2}$^\ast$} \quad
  \textbf{Jinqiao Wang\textsuperscript{1,2}} \\
  \\
  \textsuperscript{1}School of Advanced Interdisciplinary Sciences, University of Chinese Academy of Sciences \\
  \textsuperscript{2}Foundation Model Research Center, Institute of Automation, Chinese Academy of Sciences \\
  \textsuperscript{3}Tencent \quad
  \textsuperscript{4}Wuhan University \quad
  \textsuperscript{5}Southeast University \quad
  \textsuperscript{6}National University of Singapore \\
  {\footnotesize
\texttt{wangtianyue25@mails.ucas.ac.cn,
leigangqu@gmail.com,
haiyun.guo@nlpr.ia.ac.cn}
}
}

\begin{document}
\maketitle

% \blfootnote{$^\ast$ Corresponding authors. \quad
%             $^\dagger$ Project leader.}
\blfootnote{$^\ast$ Corresponding authors.}
\blfootnote{$^\dagger$ Project leader.}
            
\begin{abstract}
Long-form video understanding remains challenging for video agents due to the mismatch between query demands and evidence acquisition strategies.
Although recent planning-before-perception methods outperform query-agnostic pipelines, they often rely on a single dominant strategy, either generation-based strategy or retrieval-based strategy, limiting their ability to handle diverse query demands.
We propose \textbf{Route2Look}, a lightweight and model-agnostic framework for query-adaptive evidence acquisition in long-form video understanding.
Route2Look operates in a \textit{Route-Look-Memorize} loop with three tools: \textit{Global Browse} for holistic context, \textit{Temporal Ground} for explicit temporal cues, and \textit{Semantic Retrieve} for semantic search.
The core component is a routing policy that dynamically selects evidence acquisition tools based on the query. 
To build this policy, Route2Look adopts a two-stage design: first distilling the routing skill from differential contrastive analysis between generation-based and retrieval-based trajectories, and then applying the distilled skill with hard routing rules and continue-or-stop criteria during inference.
Experiments on challenging long-video benchmarks show that Route2Look achieves state-of-the-art performance while maintaining comparable frame efficiency across datasets and query types.
Oracle routing analysis further reveals the potential of query-adaptive evidence acquisition for future long-form video understanding.
\end{abstract}

\begin{figure*}[htbp]
    \centering
    \includegraphics[width=1.0\textwidth]{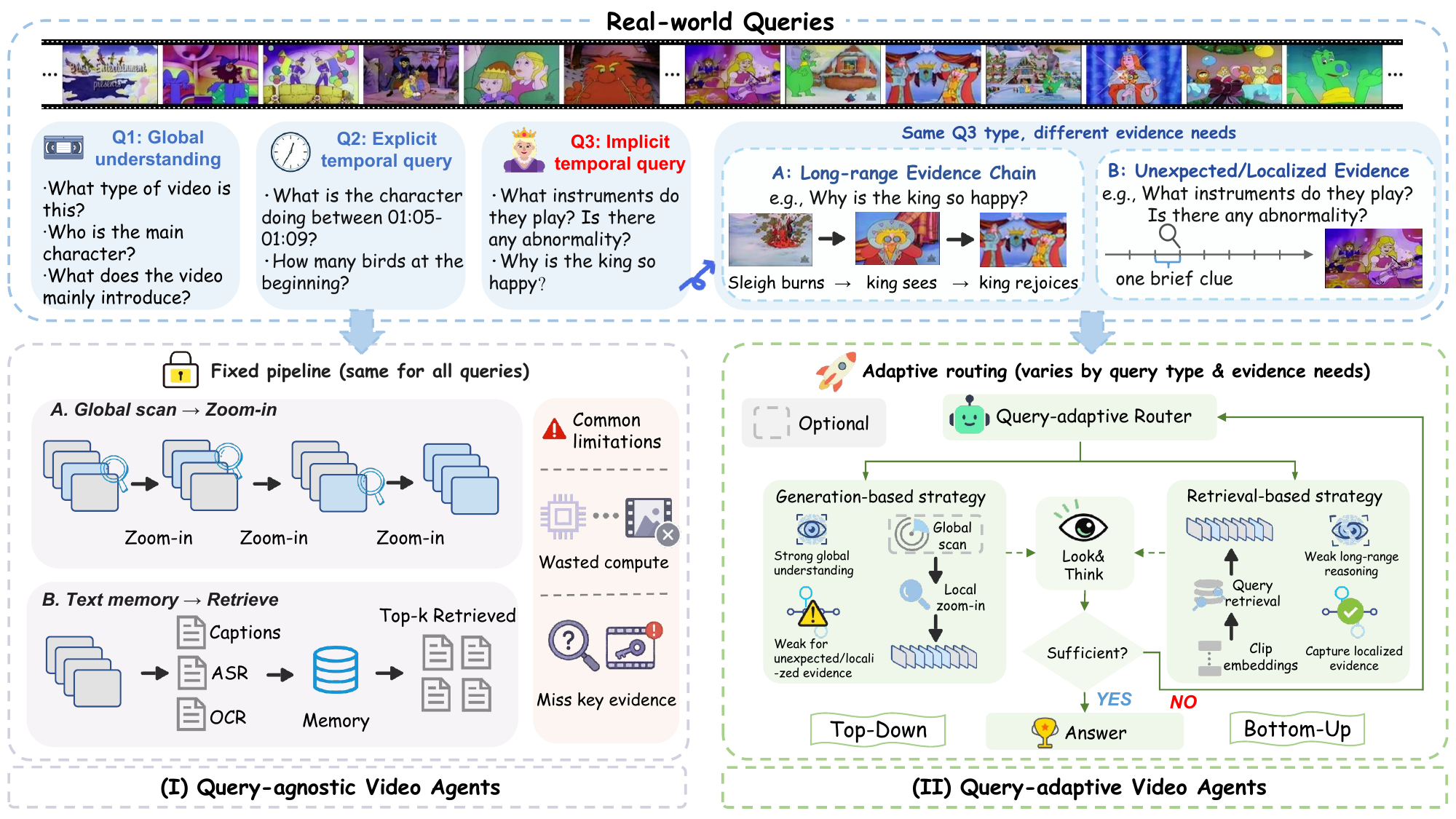}
    % \caption{Real-world long-video queries require different evidence acquisition behaviors: some need global understanding or explicit temporal grounding, while implicit temporal queries may further require either cross-segment evidence chains or brief localized clues. (I) Query-agnostic video agents apply a fixed pipeline to all queries, which may waste computation or miss key evidence. (II) In contrast, Route2Look introduces query-adaptive routing dynamically selects between the top-down generation-based strategy (following narrative logic to infer relevant intervals) and the bottom-up retrieval-based strategy (capturing salient local evidence via semantic search), ensuring efficient and accurate evidence acquisition.}
    \caption{Real-world long-video queries demand different evidence acquisition strategies. Query-agnostic agents apply a fixed pipeline to all queries, wasting computation or missing key evidence. In contrast, Route2Look introduces query-adaptive routing that dynamically selects between top-down generation-based and bottom-up retrieval-based strategies for efficient and accurate evidence acquisition.}
    \label{fig:motivation}
\end{figure*}

\section{Introduction}
Long-form video understanding has become an important frontier in multimodal intelligence, with applications in multimodal assistants~\cite{qian2025dispider, weerasinghe2024real}, embodied agents~\cite{song2025robospatial, zitkovich2023rt}, and autonomous systems~\cite{tian2024drivevlm, jiang2025survey, wei2025ad}. 
Recent progress in multimodal large language models (MLLMs)~\cite{wang2024qwen2, shu2025video, lin2026unleashing, qiang2025ver, wei2026seeing} and video agent frameworks~\cite{ma2025drvideo, yan2026symphony,yang2026videodetective,zhao2026pyvision,yang2026svagent} has enabled models to process increasingly long and complex videos. 
However, a key problem remains unsolved: the systematic mismatch between query demands and evidence acquisition strategies.

Based on observations, we identify three distinct query types in real-world scenarios as illustrated in Fig.~\ref{fig:motivation}:
(1) global understanding queries, such as summarization, main theme identification, or overall video categorization;
(2) explicit temporal queries, where the query contains timestamps, time ranges, or temporal expressions; and
(3) implicit temporal queries, a common and challenging type where the relevant evidence is not explicitly localized by the query.
For the first two types, evidence acquisition can often be guided by either global coverage or explicit temporal constraints.
The core difficulty lies in the third type: before answering \emph{what} happened, the model needs to infer \emph{where} to acquire relevant evidence.

Existing agentic systems for long-form video understanding fall into two lines. 
Query-agnostic pipelines apply a fixed pipeline (e.g., global scanning or precomputed textual representations) to all queries, wasting computation on irrelevant content~\cite{yang2025vca,zuo2026videolucy,yang2025longvt,he2025framethinker,wang2026think,yin2025videoarm,wang2024videoagent,wang2025videotree,ma2025drvideo,luo2026video,zhang2026deep}. Query-conditioned pipelines enable adaptive acquisition while still relying on a single dominant strategy for inspection, particularly for implicit temporal queries. These strategies fall into two paradigms based on how temporal intervals are selected. 
Generation-based strategies adopt a \textit{top-down} paradigm, reasoning over the query and accumulated observations to progressively generate temporal intervals~\cite{lin2026videoseek, li2026lenswalk, zhang2026eva}. They excel at inference from global context or logic flow but may miss unexpected, localized events~\cite{lin2026videoseek}.
In contrast, retrieval-based strategies adopt a \textit{bottom-up} paradigm, retrieving semantically relevant temporal intervals in parallel via clip-level embeddings~\cite{chen2025lvagent,yuan2025videoexplorer,chen2026scaling}.
They capture salient local evidence but struggle with long-range causal chains or holistic narratives.

% Existing agentic systems for long video understanding can be broadly grouped into two lines.
% The first line follows query-agnostic pipelines, either globally scanning and progressively zooming in or precomputing textual memories such as dense captions, ASR, and OCR for downstream reasoning~\cite{yang2025vca,zuo2026videolucy,yang2025longvt,he2025framethinker,wang2026think,yin2025videoarm,wang2024videoagent,wang2025videotree,ma2025drvideo,luo2026video,zhang2026deep}.
% Such fixed procedures may waste computation on irrelevant content while missing required by the current query.
% The second line introduces query-conditioned planning before detailed perception, but most methods still rely on a single dominant strategy for inspection, especially for implicit temporal queries.
% Generation-based methods follow a \textit{top-down} paradigm, progressively generating temporal intervals through reasoning~\cite{lin2026videoseek, li2026lenswalk, zhang2026eva}; they exploit global context or video logic flow but may miss unexpected or highly localized events~\cite{lin2026videoseek}.
% Retrieval-based methods follow a \textit{bottom-up} paradigm, retrieving semantically relevant segments via clip-level embeddings~\cite{chen2025lvagent,yuan2025videoexplorer,chen2026scaling}; they capture local evidence but may struggle with long-range causal chains or holistic narrative understanding.

Overall, existing methods lack the ability to adapt evidence acquisition strategy to diverse query demands. 
To quantify this gap, we conduct an oracle study on a subset of LVBench~\cite{wang2025lvbench}: if one could perfectly select between generation-based and retrieval-based strategies per query, accuracy improves from 68.5\% (best single strategy) to 81.5\%, yielding a substantial 13.0\% absolute gain. 
This strongly suggests that query-adaptive evidence acquisition is essential for long-form video understanding.

Inspired by this, we propose \textbf{Route2Look}, a framework for query-adaptive evidence acquisition in long video understanding.
It operates in a \textit{Route-Look-Memorize} loop: it first routes the query to a suitable evidence acquisition strategy, then looks into the selected video evidence, and finally memorizes verified evidence for subsequent routing.
The framework provides three tools: \textit{Global Browse} for holistic context building, \textit{Temporal Ground} for explicit temporal grounding, and \textit{Semantic Retrieve} for parallel semantic search.
To decide which tool to invoke for evidence acquisition, Route2Look learns a reusable routing skill through a two-stage design: it first distills the routing skill from differential contrastive analysis between generation-based and retrieval-based trajectories, and then applies the distilled skill with hard routing rules and continue-or-stop criteria during inference.
As a result, Route2Look provides a lightweight, model-agnostic, and transferable framework for long-form video understanding.

In summary, our contributions are threefold:
\begin{itemize}
\item We identify {query-adaptive evidence acquisition} as a core challenge in long-form video understanding, demonstrating that different query types demand distinct evidence acquisition strategies instead of a single fixed one.
\item We propose Route2Look, a two-stage framework that distills routing skill from differential contrastive analysis between retrieval-based and generation-based strategies, enabling query-adaptive strategy selection.
\item Extensive experiments on long-form video benchmarks demonstrate that Route2Look improves both accuracy and efficiency, generalizing across datasets and query types. Oracle routing results further highlight significant potential for future query-adaptive research.
\end{itemize}

\section{Related Works}
\subsection{Video Agentic Models}

Recent advances in MLLMs have promoted agentic systems for long-form video understanding.
Unlike single-pass video processing, video agents interact with videos through iterative perception, reasoning, and tool use, making them suitable for temporally extended scenarios.
Existing methods can be broadly divided into two lines.
The first line follows \textit{query-agnostic pipelines}: some methods perform global scanning followed by progressive zoom-in~\cite{yang2025vca,zuo2026videolucy,yang2025longvt,he2025framethinker,yin2025videoarm}, while others precompute textual memories such as dense captions, ASR, or OCR for downstream retrieval and reasoning~\cite{wang2024videoagent,wang2025videotree,ma2025drvideo,pang2025mr,tian2025ego,gao2025agentic,luo2026video,zhang2026deep,lian2026verbatim}.
Although effective, these methods apply a fixed processing procedure to different queries, which may waste computation on irrelevant content or miss query-specific evidence.
The second line introduces \textit{query-conditioned planning} before detailed perception~\cite{chen2025lvagent,yuan2025videoexplorer,wang2025active,lin2026videoseek,li2026lenswalk,zhang2026eva}.
These methods improve adaptivity by using the query to guide inspection, but they usually rely on a single dominant evidence acquisition strategy, such as global browsing with local refinement or retrieval with verification.

\subsection{Automatic Skill Self-Evolution}

Automatic skill self-evolution has become an important direction for improving the adaptability of language agents~\cite{wang2023voyager,shinn2023reflexion,zhao2024expel,zheng2025skillweaver,yang2026autoskill,alzubi2026evoskill,ni2026trace2skill}.
Instead of relying on manually designed prompts or fixed tool-use policies, these methods extract generalizable skills from trajectories, interactions, or execution feedback, and reuse them in future tasks.
Prior work studies skill generation through exploration~\cite{zheng2025skillweaver}, skill extraction from user-agent interactions~\cite{yang2026autoskill}, structured skill memory~\cite{jiang2026xskill,zhou2026memento}, and failure-driven behavior refinement~\cite{alzubi2026evoskill}.
ExpeL~\cite{zhao2024expel} extracts general insights from success--failure experiences, while Trace2Skill~\cite{ni2026trace2skill} distills trajectory-local lessons into transferable declarative skills through parallel consolidation.
Sharing this trajectory-to-skill paradigm, our method extends it to long-form video understanding through strategy-level differential contrastive analysis: it compares generation-based and retrieval-based trajectories for the same query--video pair using an accuracy-first and efficiency-second criterion, and distills their relative preference into a routing skill for query-adaptive evidence acquisition.

\begin{figure*}[htbp]
    \centering
    \includegraphics[width=1.0\textwidth]{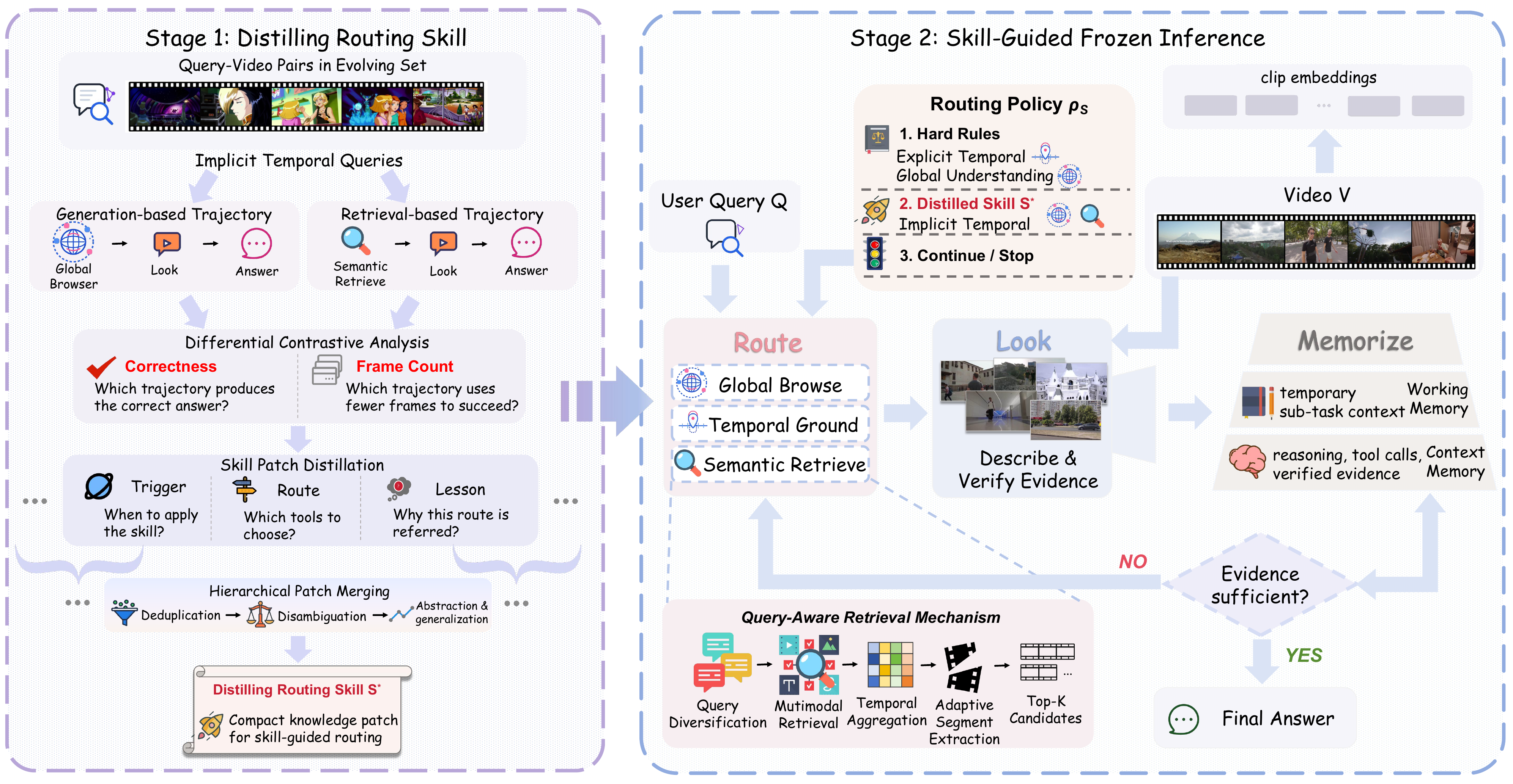}
    \caption{\textbf{Overview of Route2Look.}
    The agent follows a Route-Look-Memorize loop for iterative evidence acquisition.
    In Stage 1, Route2Look compares generation-based and retrieval-based trajectories for the same query-video pairs in an evolving set, then distills the resulting strategy preferences into a reusable routing skill.
    In Stage 2, the distilled routing skill is combined with hard routing rules and stopping criteria to guide query-adaptive evidence acquisition on the test set without updating the backbone model.}
    \label{fig:two_stage}
\end{figure*}

% \begin{figure*}[htbp]
%     \centering
%     Previous alternative figure omitted from the arXiv source package.
%     \caption{\textbf{Overview of Route2Look.}
%     The agent operates in a Route-Look-Memorize loop for iterative evidence acquisition.
%     In Stage 1, Route2Look compares generation-based and retrieval-based trajectories under an accuracy-first efficiency criterion, and distills the resulting strategy preferences into reusable routing skills.
%     In Stage 2, the learned skill is combined with hard routing rules and continue-or-stop criteria to guide query-adaptive evidence acquisition during inference, without updating the underlying MLLM.}
%     \label{fig:two_stage}
% \end{figure*}

\section{Method}
\label{sec:method}
% In this work, we propose \textbf{Route2Look}, a query-adaptive framework for evidence acquisition in long-form video understanding, as shown in Fig.~\ref{fig:two_stage}.
% We first introduce the agentic evidence acquisition protocol, which formalizes the interaction between the agent and the video through iterative tool use (Sec.~\ref{sec:pipeline}).
% We then describe the two-stage framework that distills routing experience into reusable skill and applies the learned skill during frozen inference (Sec.~\ref{sec:two_stage}).

In this work, we propose \textbf{Route2Look}, a query-adaptive framework for evidence acquisition in long-form video understanding (Fig.~\ref{fig:two_stage}). 
We first formalize an agentic evidence acquisition protocol (Sec.~\ref{sec:pipeline}), then present a two-stage framework that distills routing experience into reusable skill and applies it during frozen inference (Sec.~\ref{sec:two_stage}).

\subsection{Agentic Evidence Acquisition Protocol}
\label{sec:pipeline}
Route2Look operates in a \textit{Route-Look-Memorize} loop. 
Given query $Q$ and video $V$ of duration $l$, the agent iteratively acquires evidence until sufficient to answer $Q$. 
At each iteration, it selects an acquisition tool, inspects visual content, updates memory and decides whether to continue. 
The agent $\pi_\theta$ remains frozen throughout.

\subsubsection{Toolkit}
The agent is equipped with three evidence acquisition tools, each designed for a distinct acquisition pattern.

\paragraph{Global Browse.}
It uniformly samples frames across the full video to provide holistic context, supporting global understanding queries or providing a coarse contextual prior before more localized inspection.

\paragraph{Temporal Ground.}
It uniformly samples frames within a definite temporal interval, specified by queries containing explicit timestamps, temporal ranges, or temporal descriptions.

\paragraph{Semantic Retrieve.}
It performs query-aware semantic search over the entire video, as detailed in Sec.~\ref{sec:retrieval}.
Rather than sequential browsing, it retrieves candidate clips in parallel by ranking multimodal embedding relevance to the query, followed by frame sampling from the retrieved candidates.

\subsubsection{Route--Look--Memorize}
\paragraph{Route.}
At iteration $t$, the agent receives the current query $Q_t$ (initialized as $Q$ and updated with memory). 
Given $Q_t$ and the available tools $\mathcal{T}$, the agent performs routing reasoning $r_t$ and selects a tool $u_t \in \mathcal{T}$ via $\pi_{\theta}$. 
For global understanding queries and explicit temporal queries, the agent applies deterministic routing rules.
For more challenging implicit temporal queries, the agent decides whether to adopt a generation-based strategy through Global Browse or a retrieval-based strategy through Semantic Retrieve.

\paragraph{Look.}
The selected tool returns sampled frames $F_t$, which are then passed to a Vision-Language Model (VLM) for detailed visual inspection.
The VLM generates descriptions and assesses whether the content provides useful evidence, filtering irrelevant segments while preserving evidence $E_t$ for the final answer.

\paragraph{Memorize.}
% To support coherent multi-step reasoning, Route2Look maintains a memory structure with two complementary components.
Two complementary memory components are maintained: \textit{Context memory} $\mathcal{M}_c$ stores the full interaction history $\{(r_i, u_i, E_i)\}_{i=1}^t$ to support coherent reasoning; \textit{Working memory} $\mathcal{M}_w$ maintains focused context for sub-tasks, and only verified evidence returns to $\mathcal{M}_c$.

After each iteration, the agent evaluates whether $\mathcal{M}_c$ contains sufficient evidence to answer $Q$. 
If yes, it terminates and generates the final answer; otherwise, it proceeds to the next iteration (limited by a maximum iteration budget $H$).

\subsubsection{Query-Aware Retrieval Mechanism}
\label{sec:retrieval}
We now detail the retrieval mechanism used by Semantic Retrieve.

\paragraph{Query Diversification.}
To improve recall, we expand the current query $Q_t$ into a set of diversified sub-queries $Q_t^i$ which include multiple reformulations, higher-level semantic abstractions, alternative perspectives, and variants that incorporate information from multiple-choice options.
This diversification increases the probability of matching relevant visual content across different semantic framings ~\cite{wang2026wiser}.

\paragraph{Multimodal Retrieval.}
The video is divided into $N_c$ non-overlapping clips $\{c_j\}_{j=1}^{N_c}$.
Each sub-query $Q_t^i \in \mathcal{Q}_t$ and each clip $c_j$ is encoded into a shared multimodal embedding space.
We compute a temporal relevance score for each sub-query:
\begin{equation}
    \mathbf{s}_i[j] = \text{cos}\big({\Phi}(Q_t^i), \Phi(c_j)\big), \quad j = 1, \dots, N_c,
\end{equation}
where $\Phi$ denotes a pre-trained multi-modal encoder, and $\text{sim}(\cdot,\cdot)$ denotes cosine similarity.

\paragraph{Temporal Evidence Aggregation.}
To combine signals from different query formulations, we apply element-wise max pooling over all relevance score curves:
\begin{equation}
    \mathbf{s}_{\text{fused}}[j] = \max_i \mathbf{s}_i[j], \quad j = 1, \dots, N_c.
\end{equation}
This fused score curve highlights temporal locations that match any diversified query formulation.

\paragraph{Adaptive Segment Extraction.}
Candidate segments are extracted using a dynamic threshold:
\begin{equation}
    \gamma_{\text{ret}} = \mu(\mathbf{s}_{\text{fused}}) + \alpha \cdot \sigma(\mathbf{s}_{\text{fused}}),
\label{eq:retrieval_threshold}
\end{equation}
where $\mu(\cdot)$ and $\sigma(\cdot)$ denote the mean and standard deviation of the fused scores, and $\alpha$ controls the sensitivity of extraction.
Peaks above $\gamma_{\text{ret}}$ are identified and expanded bilaterally to cover coherent temporal events.

\paragraph{Candidate Selection.}
The extracted segments are ranked by their average fused relevance score.
The top-$K$ segments are selected as candidate evidence and passed to the Look phase for detailed verification.

\begin{table*}[t]
\centering
\caption{Main results on three long video understanding benchmarks. \#Frames denotes the number of processed frames. Best results are in \textbf{bold}, second-best are underlined.}
\label{tab:main}
\small
\begin{tabular}{c|lcccccc}
\toprule
\multirow{2}{*}{Type} & \multirow{2}{*}{Method} 
& \multicolumn{2}{c}{LVBench} 
& \multicolumn{2}{c}{VideoMME (Long)} 
& \multicolumn{2}{c}{LongVideoBench (Long)} \\
\cmidrule(lr){3-4} \cmidrule(lr){5-6} \cmidrule(lr){7-8}
& & \#Frames$\downarrow$ & Acc$\uparrow$ & \#Frames$\downarrow$ & Acc$\uparrow$ & \#Frames$\downarrow$ & Acc$\uparrow$ \\
\midrule
% \multirow{5}{*}{VLM-based} 
\multirow{7}{*}{\makecell{VLM\\-based}}
& Qwen2.5-VL-72B~\cite{bai2025qwen25vltechnicalreport} & 768 & 47.3 & 256 & 53.2 & - & - \\
& Qwen3-VL-8B~\cite{bai2025qwen3} & 768 & 45.8 & 256 & 59.3 & - & - \\
& Qwen3.5-9B~\cite{team2026qwen3} & 768 & 60.8 & 256 & 67.3 & - & - \\
& Gemini-1.5-Pro~\cite{team2024gemini} & 3600 & 33.1 & 1233 & 67.4 & 256 & 58.6 \\
& Gemini 2.0 Flash~\cite{pichai2024introducinggemini20} & 4037 & 48.3 & 1233 & 63.0 & 256 & 45.7 \\
& GPT-4o~\cite{hurst2024gpt} & 384 & 30.8 & 384 & 65.3 & 256 & 60.9 \\
& GPT-5~\cite{singh2025openai} & 384 & 60.1 & 384 & 67.9 & 384 & 64.5 \\
\midrule
\multirow{8}{*}{\makecell{Agent\\-based}}
% \multirow{8}{*}{Agent-based} 
& VideoAgent~\cite{wang2024videoagent} & 25.5 & 29.3 & 24.6 & 46.4 & - & - \\
& VideoTree~\cite{wang2025videotree} & 103.2 & 28.8 & 98.0 & 53.1 & - & - \\
& DrVideo~\cite{ma2025drvideo} & - & - & 493.2 & 51.7 & - & - \\
& VCA~\cite{yang2025vca} & 20.0 & 41.3 & 18.1 & 54.2 & - & - \\
& Mr.Video~\cite{pang2025mr} & 8074 & 60.8 & 4932 & 61.8 & 2816 & 61.6 \\
& DVD~\cite{zhang2026deep} & 8074 & \underline{74.2} & 4932 & 67.3 & 2816 & 68.6 \\
& VideoSeek~\cite{lin2026videoseek} & 92.3 & 68.4 & 60.9 & \underline{70.1} & 29.6 & \underline{73.5} \\
& \tfc \textbf{Route2Look (Ours)} & \tfc {202.3} & \tfc \textbf{75.4} & \tfc {126.6} & \tfc \textbf{76.1} & \tfc {168.7} & \tfc \textbf{77.8} \\
\bottomrule
\end{tabular}
\end{table*}

\subsection{Two-Stage Framework: From Experience to Skill for Query-Adaptive Routing}
\label{sec:two_stage}
The evidence acquisition protocol defines how the agent acquires evidence once a tool is selected.
The remaining challenge is how to choose the optimal tool for each query.
To address this, Route2Look uses a two-stage framework: 
% Stage 1 distills routing knowledge from competing trajectories, and Stage 2 applies the distilled skill during frozen inference.
Stage 1 distills routing knowledge from competing trajectories on an evolving set $\mathcal{D}_{\text{evolve}}$; Stage 2 applies the distilled skill during frozen inference on test set $\mathcal{D}_{\text{test}}$.

\subsubsection{Routing Skill Evolution Formalization}
\label{sec:formalization}
We define a routing skill $\mathcal{S}$ as a structured, human-readable knowledge document that specifies query-adaptive routing strategies, i.e., when to use a generation-based or retrieval-based strategy.
Let $\mathcal{D}_{\text{evolve}}$ and $\mathcal{D}_{\text{test}}$ denote two disjoint query-video sets for skill evolution and evaluation, respectively.

Given a skill $\mathcal{S}$, we measure its answer accuracy on a dataset $\mathcal{D}$ as
\begin{equation}
\mathcal{P}(\mathcal{S}; \pi_{\theta}, \mathcal{D})
=
\frac{1}{|\mathcal{D}|}
\sum_{(Q,V,y^*) \in \mathcal{D}}
\mathbf{1}\big[\pi_{\theta}(Q,V;\mathcal{S}) = y^*\big],
\end{equation}
where $y^*$ denotes the ground-truth answer.
We further measure frame efficiency by the average number of inspected frames:
\begin{equation}
\mathcal{C}(\mathcal{S}; \pi_{\theta}, \mathcal{D})
=
\frac{1}{|\mathcal{D}|}
\sum_{(Q,V,y^*) \in \mathcal{D}}
c(Q,V;\mathcal{S}),
\end{equation}
where $c(Q,V;\mathcal{S})$ is the number of frames inspected when answering query $Q$ on video $V$.

The objective of skill evolution is to construct a routing skill from competing trajectories on $\mathcal{D}_{\text{evolve}}$, without updating agent parameters $\theta$:
% \begin{equation}
% \mathcal{S}^* = \operatorname{Evolve}(\mathcal{S}_0, \mathcal{D}_{\text{evolve}}; \pi_{\theta}),
% \end{equation}
% \begin{equation}
% \begin{cases}
% \mathcal{P}(\mathcal{S}^*; \pi_{\theta}, \mathcal{D}_{\text{test}}) > \mathcal{P}(\mathcal{S}_0; \pi_{\theta}, \mathcal{D}_{\text{test}}), \\
% \text{or} \\
% \mathcal{P}(\mathcal{S}^*; \pi_{\theta}, \mathcal{D}_{\text{test}}) = \mathcal{P}(\mathcal{S}_0; \pi_{\theta}, \mathcal{D}_{\text{test}}) \\
% \quad \text{and} \quad \mathcal{C}(\mathcal{S}^*; \pi_{\theta}, \mathcal{D}_{\text{test}}) < \mathcal{C}(\mathcal{S}_0; \pi_{\theta}, \mathcal{D}_{\text{test}}).
% \end{cases}
% \end{equation}
\begin{equation}
\mathcal{P}(\mathcal{S}^*; \pi_{\theta}, \mathcal{D}_{\text{test}}) > \mathcal{P}(\mathcal{S}_0; \pi_{\theta}, \mathcal{D}_{\text{test}}),
\end{equation}
or, when accuracies are equal,
\begin{equation}
\mathcal{C}(\mathcal{S}^*; \pi_{\theta}, \mathcal{D}_{\text{test}}) < \mathcal{C}(\mathcal{S}_0; \pi_{\theta}, \mathcal{D}_{\text{test}}).
\end{equation}
The initial skill $\mathcal{S}_0$ represents a single fixed strategy (either generation-based or retrieval-based). 
The goal is to learn a routing skill $\mathcal{S}^*$ that outperforms the better of these two baselines.

% In other words, $\mathcal{S}^*$ is preferred if it achieves higher accuracy than using both strategies without routing, or comparable accuracy with fewer frames.

\subsubsection{Stage 1: Distilling Routing Skill from Competing Trajectories}
\label{sec:stage1}

This stage aims to attain a generalizable routing skill $\mathcal{S^*}$ from an evolving set $\mathcal{D}_{\text{evolve}}$.

\paragraph{Dual-Trajectory Sampling.}
For each query-video pair $(Q, V) \in \mathcal{D}_{\text{evolve}}$, we independently execute two distinct agentic trajectories within the pipeline defined in Sec.~\ref{sec:pipeline}.
For each implicit temporal query in $\mathcal{D}_{\text{evolve}}$, we execute two competing trajectories: a generation-based trajectory $\tau^{\text{gen}}$ (using Global Browse) and a retrieval-based trajectory $\tau^{\text{ret}}$ (using Semantic Retrieve). 
For explicit or global queries, both trajectories follow the same deterministic routing.

\paragraph{Differential Contrastive Analysis.}
A preference pair $(\tau^+, \tau^-)$ is formed when one trajectory outperforms the other according to the accuracy-first, efficiency-second criterion: $\tau^+$ is preferred if it answers correctly while $\tau^-$ fails, or if both answer correctly but $\tau^+$ inspects fewer frames.

% \paragraph{Skill patch distillation.}
% For each preference pair, $\pi_\theta$ distills a skill patch $p_i = (\text{Trigger}_i, \text{Strategy}_i, \text{Lesson}_i)$, where Trigger describes query characteristics, Strategy recommends generation or retrieval, and Lesson provides actionable guidelines. The patch converts instance-level trajectory comparisons into reusable routing knowledge.

\paragraph{Skill Patch Distillation.}
For each preference pair, $\pi_\theta$ distills a skill patch 
$p_i = (\textit{Trigger}_i, \textit{Strategy}_i, \textit{Lesson}_i)$, 
where \textit{Trigger} describes the query characteristics, 
\textit{Strategy} recommends either generation or retrieval, 
and \textit{Lesson} provides actionable routing guidelines. 
The patch converts instance-level trajectory comparisons into reusable routing knowledge.

\paragraph{Hierarchical Patch Merging.}
The extracted patches may be redundant or conflicting. We merge them hierarchically: patches are grouped into batches and iteratively merged via the LLM $\pi_{\theta}$, which removes duplicates, resolves conflicts by favoring higher-support patches, and abstracts instance-specific observations into general principles. This process repeats until the number of patches falls below the batch size, yielding a compact routing skill $\mathcal{S}^*$.

\subsubsection{Stage 2: Skill-Guided Frozen Inference}
\label{sec:stage2}
During inference, the distilled skill $\mathcal{S}^*$ is injected into the agent's routing policy $\rho_{\mathcal{S}}$, which consists of three layers. 
\textbf{Layer 1} applies hard routing rules: Temporal Ground for explicit temporal hints and Global Browse for global understanding queries. 
\textbf{Layer 2} employs the learned skill $\mathcal{S}^*$ to guide implicit temporal queries toward either retrieval-based or generation-based strategies. 
\textbf{Layer 3} implements continue-or-stop criteria, terminating the loop when sufficient evidence is accumulated or the maximum iteration budget $H$ is reached.

\section{Experiments}
\label{sec:experiments}

% \subsection{Experimental Settings}
% \paragraph{Benchmarks.}
% We evaluate Route2Look on three representative long-form video understanding benchmarks.
% LVBench~\cite{wang2025lvbench} contains 1,549 multiple-choice questions over 103 hour-long videos, covering diverse real-world scenarios and requiring evidence aggregation across extended temporal contexts.

% VideoMME~\cite{fu2025video}. we use its long-video split, which consists of 900 questions with an average video duration of 2,466 seconds, providing a challenging testbed for understanding temporally extended and semantically rich videos.

% LongVideoBench~\cite{wu2024longvideobench}. we evaluate on the long split of its validation set, which includes 564 questions from 188 videos with subtitles, with video durations ranging from 900 to 3,600 seconds.
% % L1-Perception 擅长捕捉“局部意外事件”，L2-Relation 要求理解“证据与全局逻辑的连接”。

% Together, these benchmarks cover a broad spectrum of long-form video scenarios, enabling a comprehensive evaluation of both the accuracy and efficiency of query-adaptive evidence acquisition.

\subsection{Experimental Settings}
\paragraph{Benchmarks.}
We evaluate Route2Look on three long-form video understanding benchmarks. LVBench~\cite{wang2025lvbench} contains 1,549 multiple-choice questions over 103 hour-long videos. VideoMME~\cite{fu2025video} (long subset) has 900 questions with an average duration of 2,466 seconds. LongVideoBench~\cite{wu2024longvideobench} (long split of its validation set) includes 564 questions from 188 videos lasting 900–3,600 seconds. 
Together, these benchmarks cover a broad spectrum of long-form video scenarios, enabling a comprehensive evaluation of both the accuracy and efficiency of query-adaptive evidence acquisition.

\paragraph{Baselines.}
We compare against two categories of methods for video understanding: 
(1) VLM-based methods, covering both commercial and open-source models, such as GPT-5, GPT-4o, Gemini-1.5 Pro, Gemini 2.0 Flash, and Qwen2.5-VL-72B.
(2) Agent-based methods, such as VideoAgent, Videotree, VCA, MR.Video, DVD, VideoSeek which also use GPT-5 as the backbone. 
Most reported results are from official leaderboards or published reports.

\paragraph{Implementation Details.}
Route2Look is model-agnostic and can be instantiated with different LLM/VLM backbones.
By default, we use GPT-5 as the agent backbone and LanguageBind~\cite{zhu2024languagebind} as the multimodal retriever using 10-second clips.
We set top-$K$ to 5, the threshold parameter $\alpha$ to 0.5, and the maximum number of agent iterations to 10.
For each tool call, videos are sampled at 2 FPS with at most 50 frames due to API input limits.
The evolving set contains 500 stratified samples from CG-Bench~\cite{chen2025cg}, and the merge batch size for hierarchical patch merging is set to $B=32$.  
Hyperparameter analysis is provided in Appendix~\ref{sec:hyperparameter}.

% \subsection{Main Results}
% Table~\ref{tab:main} presents the comparison results across three long-form video understanding benchmarks. Route2Look establishes new state-of-the-art performance on all benchmarks while maintaining competitive frame efficiency. 
% On LVBench, Route2Look achieves 75.4\% accuracy using 202.3 frames, significantly outperforming the previous SOTA DVD (74.2\% with 8,074 frames) by 1.2\% while using only 2.5\% of its frame budget. 
% On VideoMME, our method achieves 76.1\% with 126.6 frames, improving over GPT-5 (67.9\% with 384 frames) by 8.2\% and outperforming VideoSeek (70.1\%) by 6.0\%. 
% On LongVideoBench, Route2Look attains 77.8\% with 168.7 frames, surpassing DVD (68.6\%) by 9.2\% and VideoSeek (73.5\%) by 4.3\%. 
% Notably, these consistent gains are achieved with a single routing skill distilled from an external evolution set, without any per-benchmark adaptation, confirming that the learned routing captures generalizable patterns rather than dataset-specific heuristics.
% Across all three benchmarks, Route2Look consistently improves over its base VLM GPT-5 while reducing frame usage by over 50\%, demonstrating that query-adaptive routing between generation-based and retrieval-based strategies effectively allocates visual computation to query-relevant evidence, enabling state-of-the-art accuracy at a practical computational cost.

\subsection{Main Results}
Table~\ref{tab:main} shows that Route2Look achieves state-of-the-art performance on all benchmarks while maintaining comparable frame efficiency.
On LVBench, it reaches 75.4\% accuracy with 202.3 frames, outperforming DVD by 1.2\% while using only 2.5\% of its frame budget.
On VideoMME, Route2Look achieves 76.1\% with 126.6 frames, surpassing VideoSeek by 6.0\%.
On LongVideoBench, it obtains 77.8\% with 168.7 frames, improving over DVD and VideoSeek by 9.2\% and 4.3\%, respectively.
These consistent gains are achieved with the routing skill distilled from an external evolution set, showing that Route2Look learns generalizable routing patterns and improves the base model GPT-5 with substantially lower frame usage.

\begin{figure}[t]
    \centering
    \includegraphics[width=0.49\textwidth]{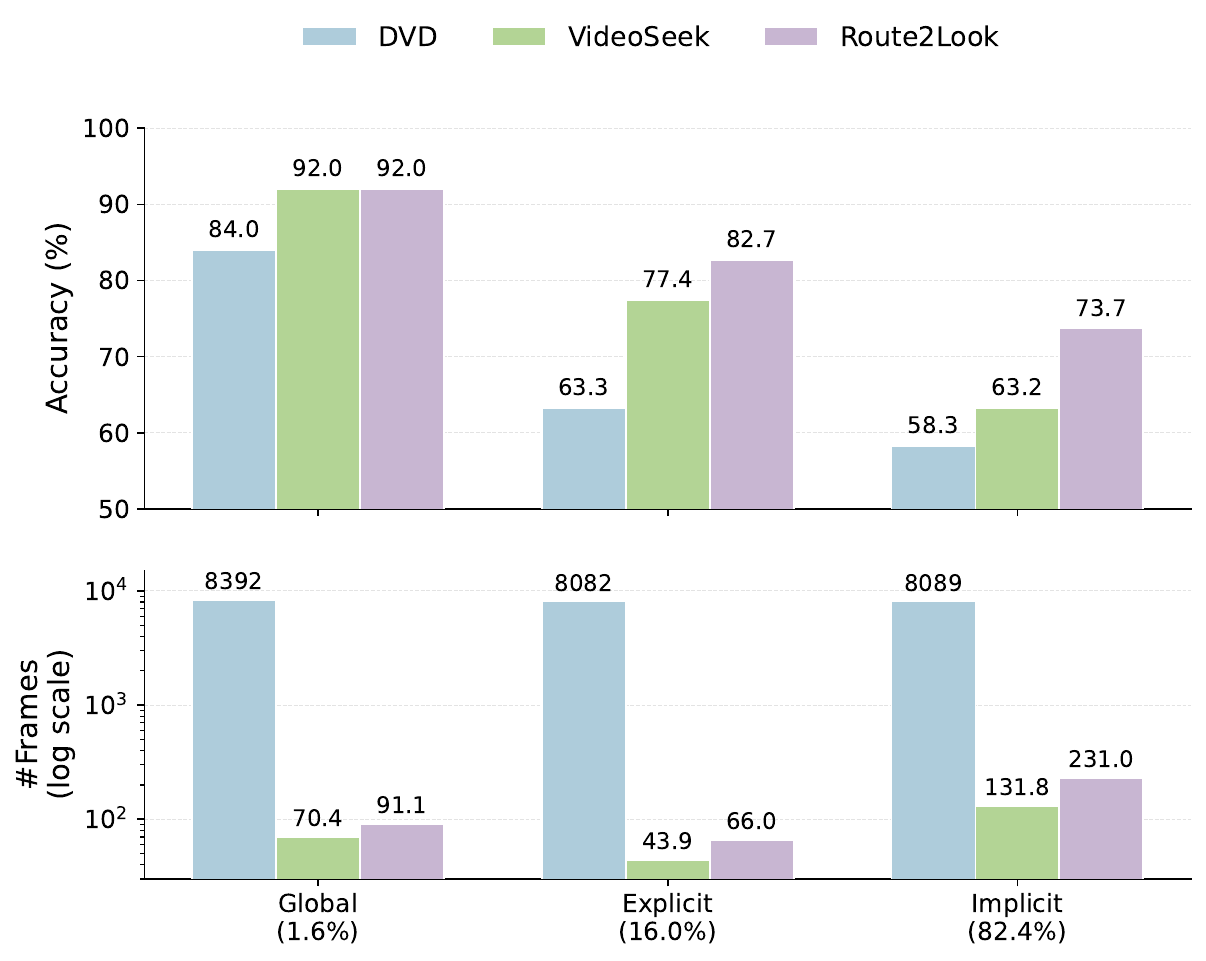}
    \caption{
    Performance breakdown by query type on LVBench.
    The upper panel shows accuracy, and the lower panel shows frame usage in log scale.
    Query-type ratios are shown in parentheses on the x-axis.
    % Route2Look achieves the best accuracy across all query types, especially on implicit temporal queries.
    }
    \label{fig:query_type}
\end{figure}

% \subsection{Performance on Different Query Types}
% \label{sec:query_type}
% We further analyze performance by query type in Fig.~\ref{fig:query_type}, including \textit{global understanding}, \textit{explicit temporal}, and \textit{implicit temporal} queries.
% For a fair comparison, DVD and VideoSeek are reproduced with GPT-5 as the same backbone.
% The three query types account for 1.6\%, 16.0\%, and 82.4\% of LVBench, respectively, indicating that implicit temporal queries are the dominant query type.
% Route2Look achieves 92.0\% accuracy on global understanding queries, matching VideoSeek and surpassing DVD by 8.0 points.
% On explicit temporal queries, it improves over VideoSeek from 77.4\% to 82.7\%, while using only 66.0 frames on average.
% The most significant gain appears on implicit temporal queries, where Route2Look reaches 73.7\% accuracy, outperforming VideoSeek by 10.5 points and DVD by 15.4 points.
% This trend shows that implicit temporal queries are not only the most common category but also the most challenging one.
% These results demonstrate that Route2Look adapts well across different query types, achieving the best accuracy while maintaining competitive frame efficiency.
% The benefit is particularly pronounced for implicit temporal queries, which further validates the effectiveness of query-adaptive evidence acquisition and shows that Route2Look can better handle diverse evidence needs in long-form video understanding.

\subsection{Performance on Different Query Types}
\label{sec:query_type}
We further analyze performance by query type in Fig.~\ref{fig:query_type}, including \textit{global understanding}, \textit{explicit temporal}, and \textit{implicit temporal} queries.
We reproduce two strong baselines, DVD and VideoSeek, with GPT-5 as the same backbone for a fair comparison.
The three query types account for 1.6\%, 16.0\%, and 82.4\% of LVBench, respectively, showing that implicit temporal queries dominate the benchmark.
Route2Look achieves 92.0\% accuracy on global understanding queries, matching VideoSeek and surpassing DVD by 8.0\%.
For explicit temporal queries, it improves over VideoSeek from 77.4\% to 82.7\% with only 66.0 frames on average.
The largest gain appears on implicit temporal queries, where Route2Look reaches 73.7\% accuracy, outperforming VideoSeek by 10.5\% and DVD by 15.4\%.
Overall, Route2Look achieves the best accuracy with competitive frame efficiency across query types, especially on the dominant and challenging implicit temporal queries.

% \begin{table*}[t]
% \centering
% \caption{Ablation studies on LVBench with breakdown by query type.}
% \label{tab:ablations_breakdown}
% \setlength{\tabcolsep}{3.5pt}
% \small
% \begin{tabular}{cccccc|cc|cc|cc}
% \toprule
% \multirow{2}{*}{Gen.} & \multirow{2}{*}{Ret.} & \multirow{2}{*}{Routing} & \multirow{2}{*}{Diff. Con.} 
% & \multirow{2}{*}{Acc} & \multirow{2}{*}{\#Frames}
% & \multicolumn{2}{c|}{Global Understanding}
% & \multicolumn{2}{c|}{Explicit Temporal}
% & \multicolumn{2}{c}{Implicit Temporal} \\
% \cmidrule(lr){7-8} \cmidrule(lr){9-10} \cmidrule(lr){11-12}
% & & & & & & Acc & \#Frames & Acc & \#Frames & Acc & \#Frames \\
% \midrule
% \checkmark &            & --      & --         & 63.5 & 107.5 
% & -- & -- & 77.1 & 52.9 & 60.6 & 119.1 \\
%            & \checkmark & --      & --         & 68.5 & 290.7 
% & -- & -- & 74.3 & 94.6 & 67.3 & 332.3 \\
% \checkmark & \checkmark & Random  & \ding{55}  & 65.8 & 196.8 
% & -- & -- & xx.x & xx.x & xx.x & xx.x \\
% \checkmark & \checkmark & Learned & \ding{55}  & xx.x & xx.x 
% & -- & -- & xx.x & xx.x & xx.x & xx.x \\
% \checkmark & \checkmark & Learned & \checkmark & \textbf{70.0} & 245.1 
% & xx.x & xx.x & xx.x & xx.x & xx.x & xx.x \\
% \checkmark & \checkmark & Oracle  & --         & 81.5 & 131.8 
% & xx.x & xx.x & xx.x & xx.x & xx.x & xx.x \\
% \bottomrule
% \end{tabular}
% \end{table*}

\begin{figure*}[htbp]
    \centering
    \includegraphics[width=1.0\textwidth]{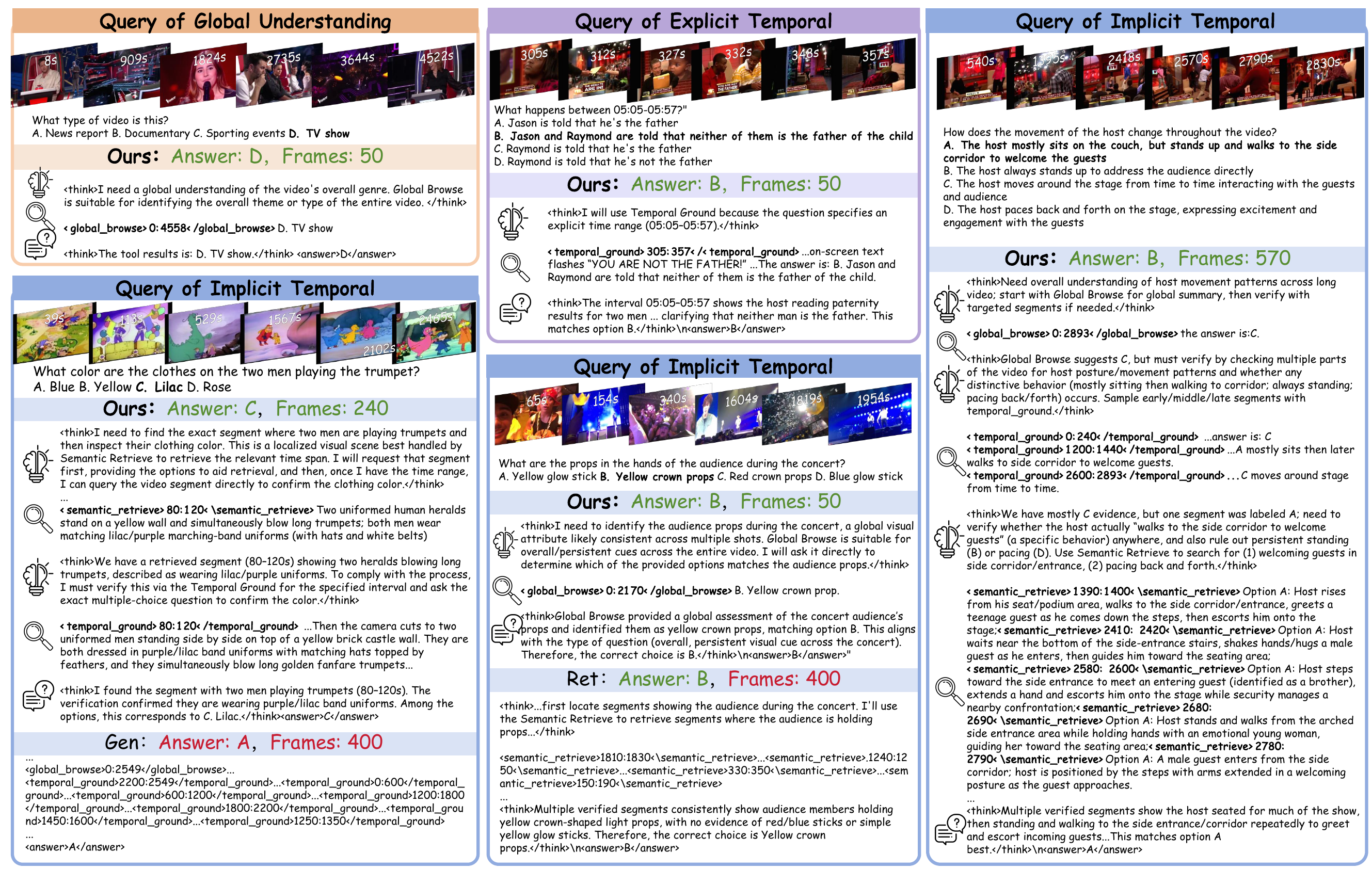}
    % \caption{Case study of Route2Look on different query types.}
    \caption{
    Case studies across different query types.
    Global understanding and explicit temporal queries are handled with low frame usage, while implicit temporal queries require adaptive evidence acquisition.
    Route2Look selects {Semantic Retrieve} for localized evidence, {Global Browse} for persistent global attributes, and multiple tools for complex reasoning, showing flexible routing across and within queries.
    }
    \label{fig:main_qualitative}
\end{figure*}

\subsection{Ablation Studies}
% We conduct ablation studies on a stratified subset of 200 samples from LVBench, which closely matches the domain composition and accuracy trends of the full dataset (see Appendix~\ref{sec:subset_analysis} for details).
We conduct ablations on a 200-sample stratified subset of LVBench that preserves the domain composition and accuracy trends of the full dataset. More details are provided in Appendix~\ref{sec:subset_analysis}.

\begin{table}[t]
    \centering
    \caption{Ablation studies on core components on LVBench.
    ``Gen.'' and ``Ret.'' denote generation-based and retrieval-based
    strategies; ``Diff. Con.'' denotes differential contrastive analysis.}
    \label{tab:ablations}

    \setlength{\tabcolsep}{4pt}
    \resizebox{\columnwidth}{!}{%
        \begin{tabular}{cccccc}
            \toprule
            Gen. & Ret. & Routing & Diff. Con. & Acc & \#Frames \\
            \midrule
            \checkmark &            & --        & --         & 63.5 & 107.5 \\
                       & \checkmark & --        & --         & 68.5 & 290.7 \\
            \checkmark & \checkmark & Random    & --         & 65.8 & 196.8 \\
            \checkmark & \checkmark & Heuristic & --         & 66.5 & 255.4 \\
            \checkmark & \checkmark & Learned   & \ding{55}  & 61.5 & 238.6 \\
            \checkmark & \checkmark & Learned   & \checkmark & \textbf{70.0} & 244.9 \\
            \checkmark & \checkmark & Oracle    & --         & 81.5 & 131.8 \\
            \bottomrule
        \end{tabular}%
    }
\end{table}

% \paragraph{Effectiveness of Core Components.}
% Since global and explicit temporal queries are handled by deterministic rules, this ablation focuses on implicit temporal queries where routing matters.
% We compare Route2Look with \textit{Generation-only}, \textit{Retrieval-only}, \textit{Random routing}, \textit{w/o differential contrastive analysis}, and \textit{Oracle}.
% The \textit{w/o differential contrastive analysis} variant summarizes each trajectory independently without explicitly comparing competing strategies, while \textit{Oracle} selects the better strategy per sample using ground-truth answers.
% As shown in Table~\ref{tab:ablations}, retrieval-only outperforms generation-only by 5.0\%, but uses 2.7$\times$ more frames, suggesting that retrieval better locates local evidence but requires more verification.
% Random routing performs between the two single-strategy variants, confirming that naive strategy mixing cannot fully exploit their complementarity.
% Removing differential contrastive analysis drops accuracy from 70.0\% to 61.5\%, showing that it is crucial for extracting useful routing preferences.
% Finally, Oracle achieves 81.5\% with only 131.8 frames, revealing substantial room for future query-adaptive routing.

\paragraph{Effectiveness of Core Components.}
Since global and explicit temporal queries are handled by deterministic rules, this ablation focuses on implicit temporal queries where routing is most critical.
We compare Route2Look with {Generation-only}, {Retrieval-only}, {Random routing}, {Heuristic routing}, {w/o differential contrastive analysis}, and {Oracle}.
As for Heuristic routing, we implement a non-learned keyword-based router that uses explicit surface cues in the query. 
For {Heuristic routing}, we implement a non-learned keyword-based router using explicit surface cues in the query.
Cues such as ``why,'' ``relationship,'' and ``throughout the video'' are routed to Global Browse, while cues such as ``what color,'' ``how many,'' and ``what is written'' are routed to Semantic Retrieve.
The {w/o differential contrastive analysis} variant summarizes each trajectory independently, while {Oracle} selects the better strategy per sample using ground-truth answers.
As shown in Table~\ref{tab:ablations}, {Retrieval-only} outperforms {Generation-only} by 5.0\% but uses 2.7$\times$ more frames, indicating a trade-off between local evidence recall and verification cost.
{Random routing} performs between the two single-strategy variants, showing that naive mixing is insufficient.
{Heuristic routing} falls 3.5 percentage points below Route2Look while consuming more frames, suggesting that the distilled skill captures nuanced routing conditions that cannot be reliably represented by surface-level keyword matching.
Removing differential contrastive analysis drops accuracy from 70.0\% to 61.5\%, confirming the importance of explicit trajectory comparison.
Notably, {Oracle} reaches 81.5\%, revealing substantial room for future query-adaptive routing.

\paragraph{Compatibility with Various Modules.}
We replace the default GPT-5 backbone in the agentic workflow with alternative LLMs and VLMs, as shown in Table~\ref{tab:backbone}.
When fixing GPT-5 as the LLM and varying the VLM for visual inspection, the accuracy steadily improves from 41.0\% with Qwen2.5-VL to 54.5\% with Qwen3.5-VL, indicating that stronger visual inspection modules directly benefit evidence verification.
When fixing GPT-5 as the VLM and varying the LLM, GPT-5 achieves 70.0\% accuracy, substantially outperforming GPT-4o and GPT-4.1 while using fewer frames.
This suggests that stronger reasoning capability improves routing and stopping decisions, leading to higher accuracy with fewer frames.

% \begin{table}[t]
% \centering
% \caption{Compatibility with different LLM and VLM backbones on LVBench.}
% \label{tab:backbone}
% % \small
% \begin{tabular}{lccc}
% \toprule
% LLM & VLM & Acc & \#Frames\\
% \midrule
% GPT-5 & Qwen2.5-VL & {41.0} & 317.7 \\
% GPT-5 & Qwen3-VL & {47.5} & 116.8 \\
% GPT-5 & Qwen3.5-VL & {54.5} & 156.2 \\
% % GPT-4o & GPT-5 & {46.0} & 171.9 \\
% GPT-4o & GPT-5 & {45.5} & 300.8 \\
% GPT-4.1 & GPT-5 & {58.0} & 282.1 \\
% GPT-5 & GPT-5 & {70.0} & 244.9 \\
% \bottomrule
% \end{tabular}
% \end{table}

\begin{table}[t]
    \centering
    \caption{Compatibility with different LLM and VLM backbones on LVBench.}
    \label{tab:backbone}

    \begin{tabular*}{\columnwidth}{
        @{\extracolsep{\fill}} lccc @{}
    }
        \toprule
        LLM & VLM & Acc & \#Frames \\
        \midrule
        GPT-5   & Qwen2.5-VL & 41.0 & 317.7 \\
        GPT-5   & Qwen3-VL   & 47.5 & 116.8 \\
        GPT-5   & Qwen3.5-VL & 54.5 & 156.2 \\
        GPT-4o  & GPT-5      & 45.5 & 300.8 \\
        GPT-4.1 & GPT-5      & 58.0 & 282.1 \\
        GPT-5   & GPT-5      & \textbf{70.0} & 244.9 \\
        \bottomrule
    \end{tabular*}
\end{table}

\subsection{Case Study}
\label{sec:case_study}
Figure~\ref{fig:main_qualitative} shows representative cases across different query types.
For global understanding and explicit temporal queries, hard routing rules invoke {Global Browse} or {Temporal Ground} and obtain correct answers with as few as 50 frames.
For implicit temporal queries, Route2Look adapts its strategy to the evidence need: it uses {Semantic Retrieve} for localized evidence, {Global Browse} for persistent global attributes, and multiple tools for complex movement reasoning.
These cases show that implicit queries do not favor a fixed strategy, and Route2Look can flexibly route among strategies across different queries and within a single query.

\section{Conclusion}
\label{sec:conclusion}
We present {Route2Look}, a lightweight and model-agnostic framework for query-adaptive evidence acquisition in long-form video understanding.
Motivated by the mismatch between query demands and fixed evidence acquisition strategies, Route2Look routes before looking and dynamically selects among three evidence acquisition tools.
It learns reusable routing skills from contrastive generation-based and retrieval-based trajectories, enabling adaptive inference without updating the backbone model.
Experiments on three long-video benchmarks demonstrate state-of-the-art accuracy with comparable frame efficiency.
Further analyses show that the gains are most pronounced on implicit temporal queries, confirming the importance of adapting evidence acquisition to query-specific evidence needs.

\section*{Limitations}
\label{sec:limitations}
Route2Look provides a first step toward query-adaptive evidence acquisition, and several limitations remain.
First, the routing skill is distilled from a finite evolution set, leaving room to close the gap between learned and oracle routing through larger data or better skill optimization.
Second, Route2Look still requires moderate frame verification, especially when Semantic Retrieve returns multiple candidates. Future work can explore more selective verification policies to further reduce frame usage while maintaining or improving accuracy.
Third, the current framework mainly focuses on visual evidence, while audio cues may provide complementary signals for long-form video understanding.
These limitations together indicate broad opportunities for future adaptive routing research.
By revealing the importance and room of query-adaptive strategy selection, Route2Look lays a foundation for building more accurate, efficient, and multimodally grounded long-video agents.

\section*{Acknowledgements}
This work was supported in part by the New Generation Artificial Intelligence-National Science and Technology Major Project (No. 2025ZD0123404) and the Beijing Natural Science Foundation (No. L252035).

\bibliography{custom}

\clearpage
\appendix

\section{Appendix}
\label{sec:appendix}

\subsection{Hyperparameter Analysis}
\label{sec:hyperparameter}
We analyze three key hyperparameters in Table~\ref{tab:hyperparameter}: the evolution set size $|\mathcal{D}_{\text{evolve}}|$, the merge batch size $B$, and the number of retrieved segments $K$. 

\paragraph{Evolution set size.} 
Increasing the evolution set from 200 to 500 samples yields a notable improvement from 67.0\% to 70.0\%, indicating that more diverse training queries help distill a more robust routing skill. 
This trend suggests that further scaling the evolution data may continue to improve routing quality.

\paragraph{Merge batch size.}
With 200 samples, batch size 32 achieves the best accuracy (67.0\%), while extremely small and overly large batches degrade performance. 
A similar trend holds with 500 samples, where increasing batch size from 32 to 200 causes a drop from 70.0\% to 64.0\%. 
These results indicate that merging too many patches at once loses fine-grained distinctions, while too few patches per batch fail to resolve conflicts effectively. 
A moderate batch size $B=32$ consistently yields the best trade-off.

\paragraph{Number of retrieved segments.}
Decreasing $K$ from 5 to 3 substantially reduces frame usage from 249.5 to 171.7, but at the cost of a notable accuracy drop from 67.0\% to 61.5\%. 
This indicates that retrieving fewer candidates improves efficiency but risks missing relevant evidence, and $K=5$ provides a favorable balance between coverage and cost.

% Overall, Route2Look performs robustly under a range of hyperparameter settings, with the default configuration (500 samples, batch size 32, $K=5$) achieving the best accuracy-efficiency trade-off.
\begin{table}[h]
    \centering
    \caption{Hyperparameter analysis on LVBench.}
    \label{tab:hyperparameter}

    \begin{tabular*}{\columnwidth}{
        @{\extracolsep{\fill}} ccccc @{}
    }
        \toprule
        $|\mathcal{D}_{\text{evolve}}|$ & $B$ & $K$ & Acc & \#Frames \\
        \midrule
        200 & 32  & 3 & 61.5 & 171.7 \\
        200 & 32  & 5 & 67.0 & 249.5 \\
        200 & 10  & 5 & 63.5 & 186.0 \\
        200 & 200 & 5 & 66.5 & 216.8 \\
        500 & 32  & 5 & \textbf{70.0} & 244.9 \\
        500 & 200 & 5 & 64.0 & 253.7 \\
        \bottomrule
    \end{tabular*}
\end{table}
% \begin{table}[h]
% \centering
% \caption{Hyperparameter analysis on LVBench.}
% \label{tab:hyperparameter}
% \begin{tabular}{ccccc}
% \toprule
% $|\mathcal{D}_{\text{evolve}}|$ & $B$ & $K$ & Acc & \#Frames \\
% \midrule
% 200 & 32 & 3 & 61.5 & 171.7 \\
% 200 & 32 & 5 & 67.0 & 249.5 \\
% 200 & 10 & 5 & 63.5 & 186.0 \\
% 200 & 200 & 5 & 66.5 & 216.8 \\
% 500 & 32 & 5 & 70.0 & 244.9 \\
% 500 & 200 & 5 & 64.0 & 253.7 \\
% \bottomrule
% \end{tabular}
% \end{table}

\subsection{Efficiency Analysis}
\label{sec:efficiency_analysis}
We further analyze the efficiency of Route2Look on LVBench.
We report the average frame usage and token consumption of the GPT-5 base model and Route2Look in Table~\ref{tab:tokens}.
Compared with the GPT-5 base model, Route2Look uses substantially fewer frames and fewer tokens, while achieving much higher accuracy.
Specifically, Route2Look reduces the average number of inspected frames from 384.0 to 202.3, corresponding to a 47.3\% reduction, and decreases token consumption from 83K to 72K.
Meanwhile, the accuracy improves from 60.1\% to 75.4\%.
These results demonstrate that query-adaptive evidence acquisition enables the agent to focus on more relevant video segments, leading to a better accuracy-efficiency trade-off.

\begin{table}[t]
    \centering
    \caption{Performance comparison of different methods.}
    \label{tab:tokens}

    \begin{tabular*}{\columnwidth}{
        @{\extracolsep{\fill}} lccc @{}
    }
        \toprule
        Method & \#Frames & \#Tokens & Acc \\
        \midrule
        GPT-5 (Base)      & 384   & 83K & 60.1 \\
        Route2Look (Ours) & 202.3 & 72K & \textbf{75.4} \\
        \bottomrule
    \end{tabular*}
\end{table}
% \begin{table}[htbp]
% \centering
% \caption{Performance comparison of different methods.}
% \label{tab:tokens}
% \begin{tabular}{lccc}
% \toprule
% Method & \#Frames & \#Tokens & Acc \\
% \midrule
% GPT-5 (Base)  & 384  & 83K & 60.1 \\
% Route2Look (Ours) & 202.3   & 72K & 75.4 \\
% \bottomrule
% \end{tabular}
% \end{table}

\subsection{Algorithm Demonstration}
\label{sec:algorithm_demo}

Algorithm~\ref{alg:route2look_workflow} illustrates the agent workflow in Route2Look.
The entire process follows a Route-Look-Memorize loop.
Given a query $Q$ and a video $V$, the agent repeatedly performs three steps.
First, it updates the current query state $Q_t$ based on context memory $\mathcal{M}_c$ and selects a tool $u_t$ from the toolkit $\mathcal{T}$ under the routing policy $\rho_{\mathcal{S}}$.
Second, the agent initializes a temporary working memory $\mathcal{M}_w^t$ for the current tool call.
The selected tool is then invoked to obtain sampled frames $F_t$, and the VLM inspects these frames to extract verified evidence $E_t$.
Third, only the verified evidence and routing information are written back to the context memory. And the agent checks whether the accumulated evidence is sufficient to answer the query.
If so, it generates the final answer $Y$; otherwise, it continues the loop until reaching the maximum iteration budget $H$.

\begin{algorithm}[t]
\caption{Route2Look Agent Workflow}
\label{alg:route2look_workflow}
\begin{algorithmic}[1]
\Require User query $Q$; video $V$; routing policy $\rho_{\mathcal{S}}$; frozen LLM/VLM agent $\pi_{\theta}$; toolkit $\mathcal{T}$; maximum iteration budget $H$.
\Ensure Answer $Y$.
\State Initialize context memory $\mathcal{M}_c \leftarrow \emptyset$
\State Initialize final answer $Y \leftarrow \emptyset$
\For{$t = 1$ to $H$}
    \State Initialize working memory $\mathcal{M}_w^t \leftarrow \emptyset$
    % \State $Q_t \leftarrow \textsc{UpdateQuery}(Q, \mathcal{M}_c)$
    \State $(r_t,u_t,Q_t) \leftarrow \pi_{\theta}(Q, \mathcal{M}_c,\mathcal{T}, \rho_{\mathcal{S}})$
    \hfill $\triangleright$ Routing reasoning, tool selection, the current query
    \State $\mathcal{M}_w^t \leftarrow \mathcal{M}_w^t \cup \{Q_t\}$
    \hfill $\triangleright$ Temporary memory for the current tool call
    \State $(F_t, \mathcal{M}_w^t) \leftarrow \textsc{ApplyTool}(u_t, V, \mathcal{M}_w^t)$
    \hfill $\triangleright$ Acquire sampled frames
    \State $(E_t, \mathcal{M}_w^t) \leftarrow \pi_{\theta}(F_t, \mathcal{M}_w^t)$
    \hfill $\triangleright$ Inspect frames and verify evidence
    \State $\mathcal{M}_c \leftarrow \mathcal{M}_c \cup \{\langle r_t,u_t,E_t\rangle\}$
    \hfill $\triangleright$ Update context memory
    \If{$\textsc{Sufficient}(Q, \mathcal{M}_c)$}
        \State $Y \leftarrow \textsc{Answer}(Q, \mathcal{M}_c)$
        \State \textbf{break}
    \EndIf
\EndFor
\If{$Y = \emptyset$}
    \State $Y \leftarrow \textsc{Answer}(Q, \mathcal{M}_c)$
    \hfill $\triangleright$ Forced final answer
\EndIf
\State \Return $Y$
\end{algorithmic}
\end{algorithm}

% We present abstract pseudocode for the LensWalk reason–plan–observe loop in Algorithm 1. Using the notation of Section 3, it shows how each Reasoner plan at = (ot, qt, It, ρot ) over tools O = {Scan Search, Segment Focus, Stitched Verify, Finish} is translated into one or more VLM-visible context groups, how the Observer Mo is queried on those groups, and how the resulting time-anchored evidence and subject table state Subt are threaded through the multi-turn loop.

\subsection{Subset Analysis in the Ablation Study}
\label{sec:subset_analysis}
We conduct ablation studies on a stratified subset of 200 samples from LVBench to reduce evaluation cost while preserving the benchmark distribution. 
Following the sampling protocol of~\cite{yin2025videoarm}, we sample from the joint domain--task distribution of the full set, prioritizing domain-level proportions and maintaining task-level proportions within each domain. 
Samples are drawn without replacement within each stratum, with minor rounding adjustment to satisfy the 200-sample budget.

\begin{figure}[t]
    \centering
    \includegraphics[width=0.5\textwidth]{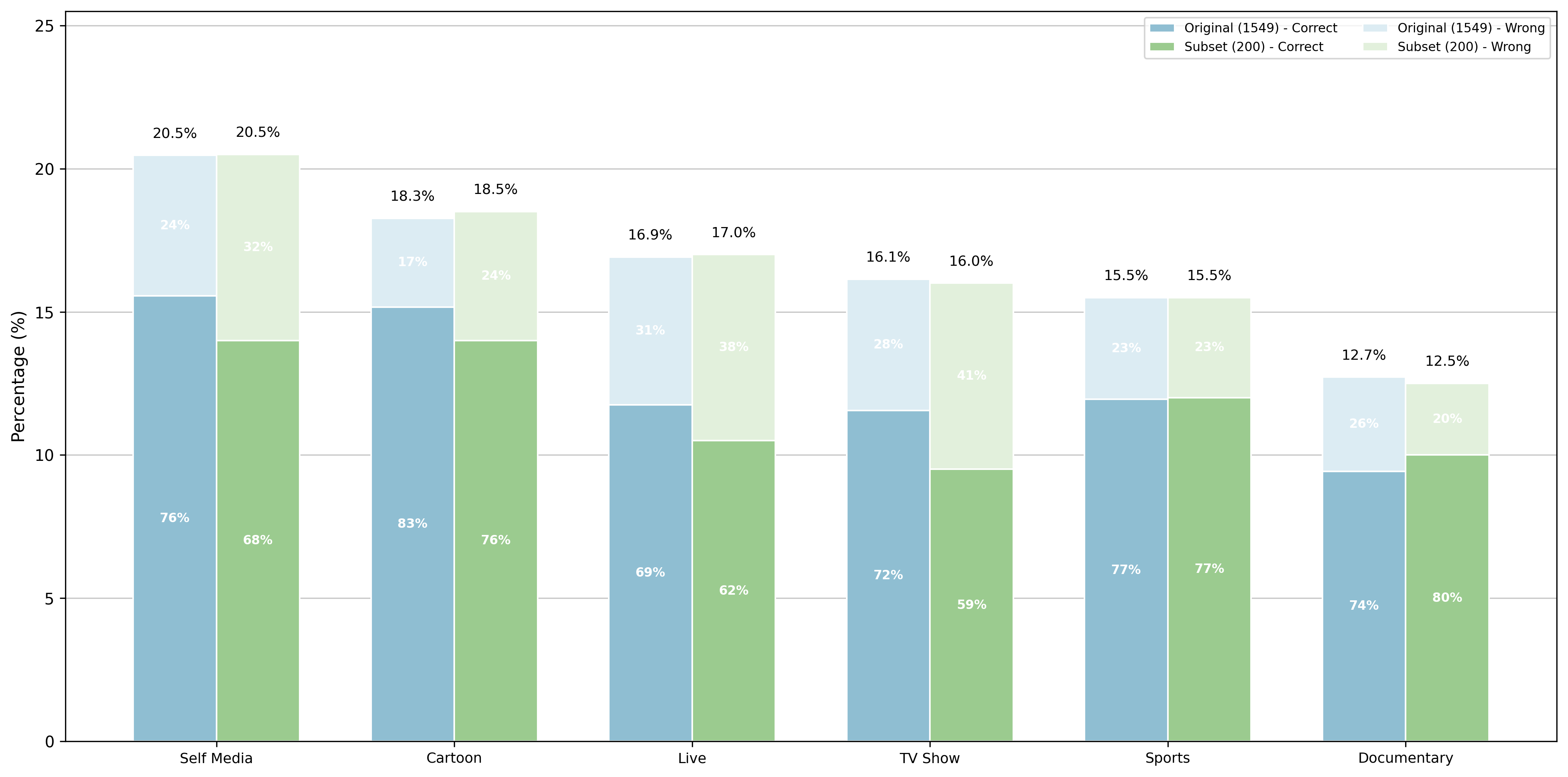}
    \caption{Effectiveness of Subset Analysis.}
    \label{fig:subset}
\end{figure}

As shown in Fig.~\ref{fig:subset}, the resulting subset closely matches the domain composition of the full LVBench set and preserves similar accuracy trends across domains. 
This indicates that the subset provides a representative and efficient testbed for controlled ablation analysis.

% This confirms that the two routes are not redundant: generation-based strategy is more suitable when the target evidence can be inferred from global context or narrative progression, whereas retrieval-based strategy is better at capturing localized or unexpected visual evidence.

\subsection{Oracle Routing Analysis}
\label{sec:oracle_analysis}
\begin{figure}[h]
    \centering
    \includegraphics[width=0.5\textwidth]{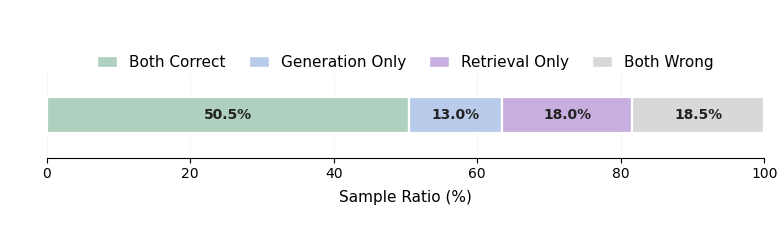}
    \caption{Oracle Routing Analysis.}
    \label{fig:oracle}
\end{figure}

We conduct an in-depth analysis of the oracle routing results, as summarized in Figure~\ref{fig:oracle}. The results show that 50.5\% of samples can be correctly answered by both strategies, indicating that a substantial portion of implicit temporal queries can be handled reliably by either top-down reasoning or bottom-up semantic retrieval.
Generation-only succeeds on 13.0\% of samples where retrieval fails, while retrieval-only succeeds on 18.0\% of samples where generation fails. By selecting the correct strategy for each sample, oracle routing achieves 81.5\% accuracy, which corresponds to the union of the two single-strategy success sets. This is substantially higher than either generation-only (63.5\%) or retrieval-only (68.5\%) performance, revealing substantial room for query-adaptive routing.
Overall, Figure~\ref{fig:oracle} provides direct empirical evidence that adaptive routing is necessary: the key challenge is not to replace one strategy with another, but to learn when each strategy should be trusted.

% Meanwhile, 18.5\% of samples remain incorrect under both strategies, suggesting that future improvements cannot rely solely on better routing. 
% These remaining failures may require stronger visual perception, more reliable temporal verification, richer multimodal evidence such as audio or subtitles, or multi-step strategy composition beyond choosing a single dominant route. 

\subsection{Generalization Beyond the Evolving Set}
\label{sec:generalization_beyond_evolving_set}

Since the routing skill is distilled from a finite evolving set, an important question is whether it generalizes beyond the observed data.
We examine this question from three perspectives: cross-benchmark transfer, held-out-domain transfer, and the effect of evolving-set scale.

\paragraph{Cross-Benchmark Generalization.}
The routing skill is distilled exclusively from CG-Bench and directly applied to LVBench, VideoMME, and LongVideoBench without re-distillation or benchmark-specific adaptation.
Its consistent gains across these independent benchmarks indicate that the learned skill captures transferable query--evidence acquisition patterns rather than dataset-specific shortcuts.

\paragraph{Held-out-Domain Evaluation.}
To more directly assess domain generalization, we conduct held-out-domain experiments on two representative video domains, as shown in Table \ref{tab:held-out-domain}.
For each target domain, we remove all corresponding samples from the CG-Bench evolving set, distill the routing skill using only the remaining domains, and evaluate it on the corresponding LVBench domain.
Specifically, we evaluate transfer from CG-Bench excluding \textit{Sports \& Exercise} to the LVBench \textit{Sport} domain, and from CG-Bench excluding \textit{Music \& TV Show} to the LVBench \textit{TV} domain.

\begin{table}[t]
    \centering
    \caption{Held-out-domain generalization on LVBench. The target domain
    is excluded from the CG-Bench evolving set during skill distillation.}
    \label{tab:held-out-domain}

    \small
    \begin{tabular*}{\columnwidth}{
        @{\extracolsep{\fill}} lccc @{}
    }
        \toprule
        Domain & Gen.-only & Ret.-only & Held-out Skill \\
        \midrule
        Sport & 71.0 & 61.3 & \textbf{74.2} \\
        TV    & 56.2 & 56.2 & \textbf{59.4} \\
        \bottomrule
    \end{tabular*}
\end{table}
% \begin{table}[t]
% \centering
% \small
% \caption{Held-out-domain generalization on LVBench. The target domain is excluded from the CG-Bench evolving set during skill distillation.}
% \label{tab:held_out_domain}
% \setlength{\tabcolsep}{6pt}
% \begin{tabular}{lccc}
% \toprule
% Domain & Gen.-only & Ret.-only & Held-out Skill \\
% \midrule
% Sport & 71.0 & 61.3 & \textbf{74.2} \\
% TV    & 56.2 & 56.2 & \textbf{59.4} \\
% \bottomrule
% \end{tabular}
% \end{table}

Despite never observing samples from the corresponding domains during skill distillation, the held-out skill outperforms both generation-only and retrieval-only baselines, demonstrating its ability to transfer to unseen video domains.

\paragraph{Effect of Evolving-Set Scale.}
Increasing the evolving set from 200 to 500 samples improves accuracy from 67.0\% to 70.0\% in Table \ref{tab:hyperparameter}, indicating that broader coverage further improves routing quality.

\subsection{Visualization of Skill}
\label{sec:skill_visualization}
To better understand what Route2Look learns during skill evolution, we visualize part of the distilled routing skill in Table~\ref{tab:skill_patches}.
The distilled skills reveal clear routing preferences.
Generation-based strategy are generally favored when the required evidence can be obtained from global context, stable visual attributes, contiguous workflows, or visually clear temporal boundaries.
For example, queries about overall video type, persistent props, structured processes, or visually defined durations often benefit from first building a coarse global timeline and then performing light local verification.
In contrast, retrieval-based strategy are preferred when the evidence is precise, sparse, transient, or lexically grounded.
Examples include exact on-screen text, brief UI changes, localized spatial relations, sparse first occurrences, or repeated similar actions that require high-recall candidate search and local verification.
These examples show that the learned skill is not a simple preference for one fixed strategy.
Instead, it captures reusable routing principles that connect query characteristics with evidence acquisition behavior.
This human-readable form also makes the routing policy interpretable: Route2Look learns when to rely on global reasoning and when to retrieve localized evidence.

\begin{table*}[t]
\centering
\small
\caption{Examples of distilled routing skill.}
\label{tab:skill_patches}
\renewcommand{\arraystretch}{1.15}
\setlength{\tabcolsep}{4pt}
\begin{tabular}{p{0.34\textwidth} p{0.16\textwidth} p{0.42\textwidth}}
\toprule
Trigger & Route & Lesson \\
\midrule

Global or persistent visual understanding, such as overall setting, participants, storyline flow, stable scene layout, persistent props, HUD/UI semantics, or category and role identification.
&
Generation-based
&
Skim the whole video to map scenes, participants, timeline, and causal flow. Use sparse uniform sampling to confirm stable cues and answer from representative frames. Escalate only when multiple scenes or configurations create ambiguity. \\

\addlinespace
Exact linguistic strings, including on-screen text, subtitles, spoken dialogue, earliest textual instances, or narration-driven entity introductions.
&
Retrieval-based
&
Use retrieval to jump to the exact segment. Densely sample adjacent frames, crop or zoom when needed, and verify exact strings across multiple frames. For earliest-instance questions, confirm that no earlier true mention exists. \\

\addlinespace
Linguistic content where exact wording is not critical, such as prominent on-screen branding, stable text layout, or a single spoken fact concentrated in one short segment.
&
Generation-based
&
Skim to capture a clear card or overlay and lightly ground the timing. Read a few clean frames or a tight transcript window. Switch to retrieval when the text is small, transient, ambiguous, or requires precise lexical distinction. \\

\addlinespace
Event-anchored localized questions tied to a clear temporal cue, such as when, after, before, beginning, first, last, explicit timestamps, or quoted lines.
&
Generation-based
&
Locate the anchor through visual cues, or transcript keywords, then bound a narrow temporal window and sample adjacent frames. Verify the simple detail, immediate outcome, or directionality without heavy retrieve--verify loops. \\

\addlinespace
Precise, localized, and time-bound visual evidence, including fine-grained spatial relations, transient UI labels, brief color or state changes, and multi-option frame-level cues.
&
Retrieval-based
&
Anchor the target moment through semantic or event-based retrieval, then densely sample a tight frame window. Inspect the relevant region, resolve viewpoint changes, track entities across cuts, and verify each claim against explicit evidence. \\

\addlinespace
Counting repeated but visually salient items or actions within a clearly bounded interval or a stable wide shot.
&
Generation-based
&
Establish the interval boundary with a global skim, then tally within the contiguous window. Use adjacent frames to stabilize visibility, broaden slightly when needed, and track objects to avoid double-counting. \\

\addlinespace
Counting or enumeration tied to a specific narrative moment, step, brief visibility window, or ordinal relation.
&
Retrieval-based
&
Retrieve the correct segment using the narrative cue or event boundary, then perform dense local verification across nearby frames. Count each visible item carefully and avoid conflating it with similar steps. \\

\addlinespace
Counting how many times a repeated, visually similar action occurs across the whole video, especially with replays, multi-angle cuts, or montages.
&
Retrieval-based
&
Retrieve candidate segments across the timeline. Verify strict start and end boundaries locally, then deduplicate cross-angle replays, montage repeats, and near-duplicate views before counting. \\

\addlinespace
Order-of-events reasoning in a structured workflow, tutorial, or contiguous process, including identifying the earliest or latest acted-on item.
&
Generation-based
&
Build a coarse global timeline through a quick skim, pin salient anchor actions, and read forward through the contiguous segment. Use brief local checks only when similar candidates remain. \\

\addlinespace
Fine-grained temporal ordering with visually similar candidates, strong distractors, or subtle action boundaries.
&
Retrieval-based
&
Anchor the referenced cue through transcript, UI, or visual semantics. Densely sample the immediate window, track entities across cuts, and verify timestamps or boundary states until the order is resolved. \\

\addlinespace
Earliest or first occurrence queries across the whole video when the target is reasonably frequent and the video is not long or cluttered.
&
Generation-based
&
Perform a global sweep to find the earliest relevant entry point, then ground and confirm the first true action or event in a tight post-cue window. Switch to retrieval if the search becomes exhaustive. \\

\addlinespace
Earliest or first occurrence in long or cluttered videos where the target event is brief, sparse, or difficult to find with high recall.
&
Retrieval-based
&
Use semantic retrieval to propose candidate moments, perform dense local verification, deduplicate near-misses, and back-check earlier windows to ensure the first true instance. \\

\addlinespace
Duration questions with visually defined boundaries, such as a continuous action, the $n$-th occurrence in a known segment, or a visually distinctive activity interval.
&
Generation-based
&
Use a global sweep to locate clear start and end boundaries, then compute elapsed time or match the closest option. Avoid sparse semantic retrieval that may miss boundary frames. \\

\bottomrule
\end{tabular}
\end{table*}

% Failure：空间上细粒度、需要长程推理的问题都依赖于VLM、LLM自身的能力，还存在挑战

% \section{suppl}
% 超参数实验、分层抽样的有效性、
% 伪代码、可视化例子(skill、success、failures)、
% Token Consumption

% efficency：avg token
% LVBench的task acc、对 temporally implicit queries选 generation 还是 retrieval 的判断准确率是多少、oracle routing 的差距在哪些 query pattern 上最大、如何分类query的、Prompt(stage1、stage2)

\subsection{Additional Case Studies}
\label{sec:additional_case_study}

\begin{figure*}[htbp]
    \centering
    \includegraphics[width=1.0\textwidth]{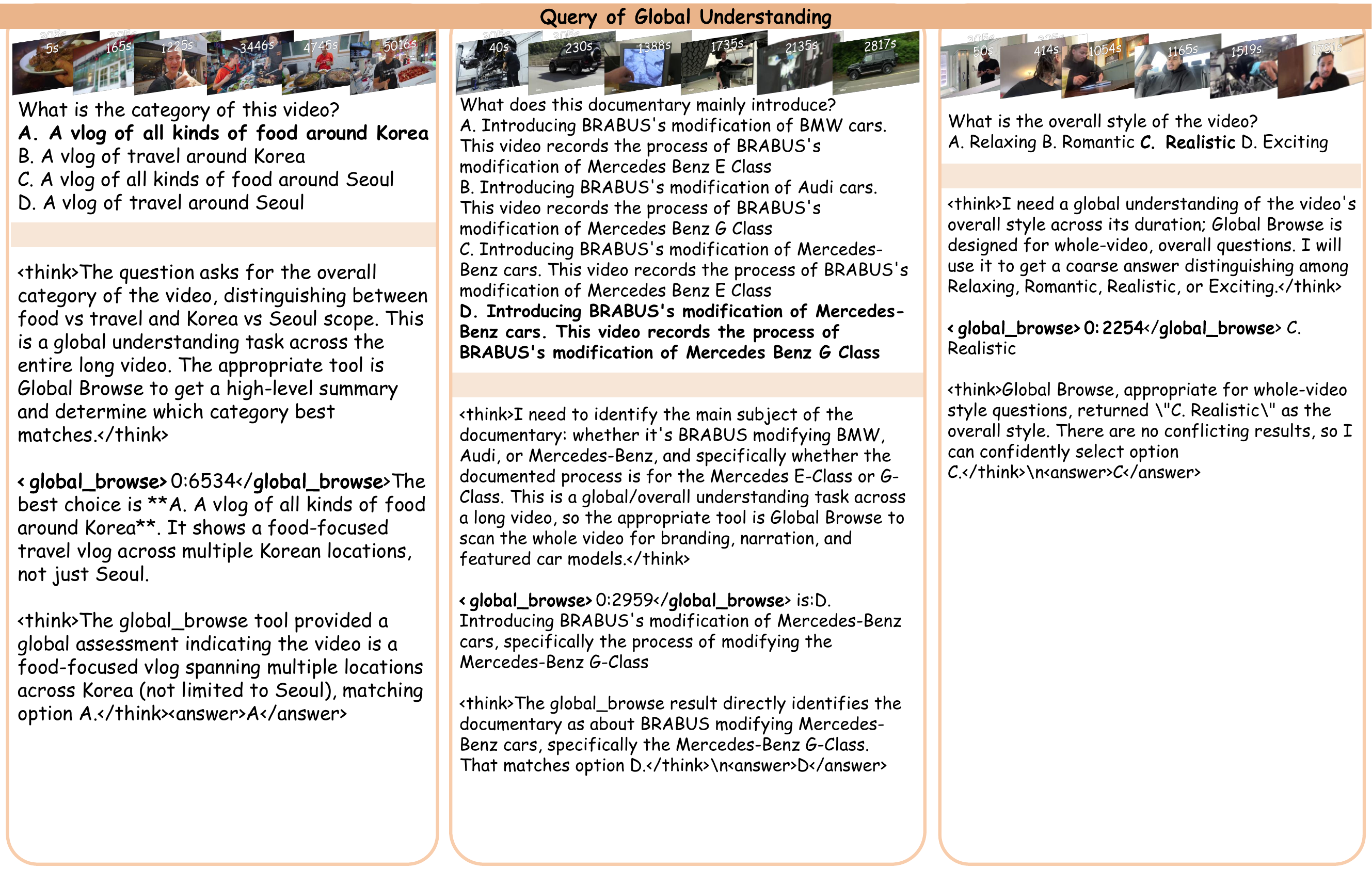}
    \caption{
    Additional case studies on global understanding queries.
    Route2Look invokes {Global Browse} to capture whole-video context and answer questions about video category, documentary topic, and overall style.
    }
    \label{fig:qualitative_global}
\end{figure*}

\begin{figure*}[htbp]
    \centering
    \includegraphics[width=1.0\textwidth]{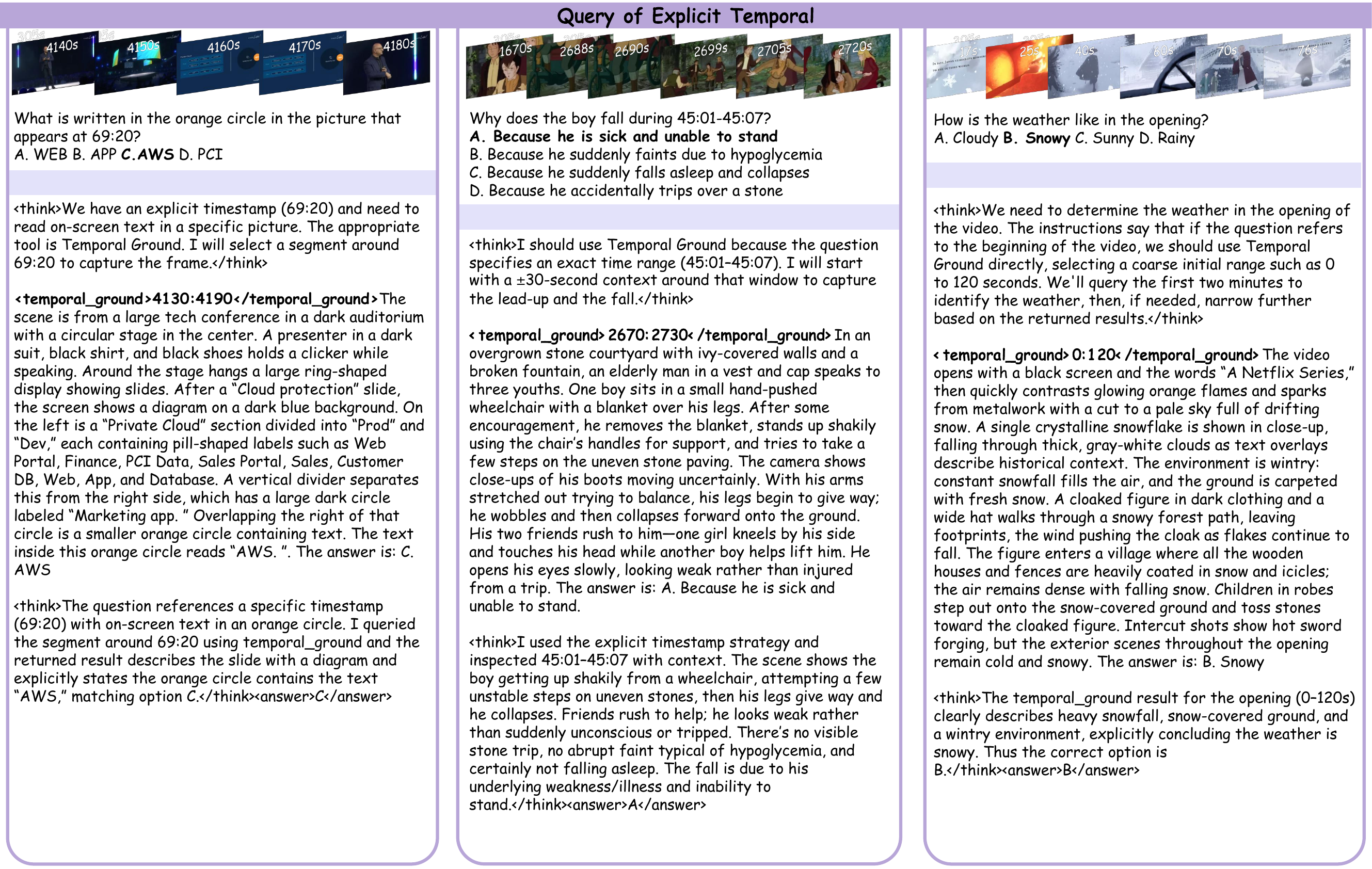}
    \caption{
    Additional case studies on explicit temporal queries.
    Route2Look invokes {Temporal Ground} when the query specifies a timestamp, time range, or temporal segment, enabling direct verification of localized evidence.
    }
    \label{fig:qualitative_explicit}
\end{figure*}

\begin{figure*}[htbp]
    \centering
    \includegraphics[width=1.0\textwidth]{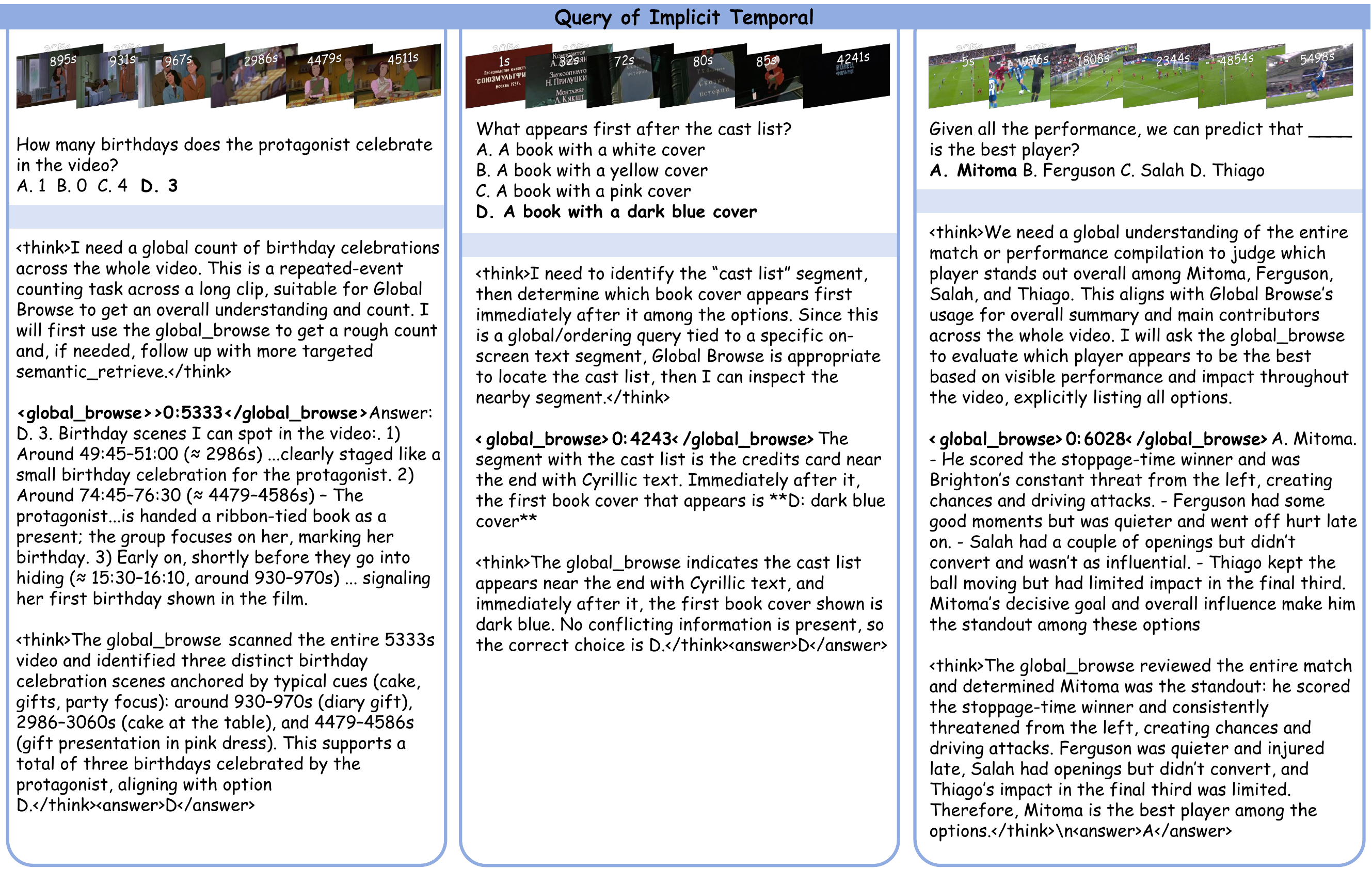}
    \caption{
    Additional case studies where implicit temporal queries favor Global Browse.
    When the query requires repeated-event counting, global ordering, or long-range reasoning, Route2Look uses {Global Browse} to aggregate evidence across the full video.
    }
    \label{fig:qualitative_implicit_global}
\end{figure*}

\begin{figure*}[htbp]
    \centering
    \includegraphics[width=1.0\textwidth]{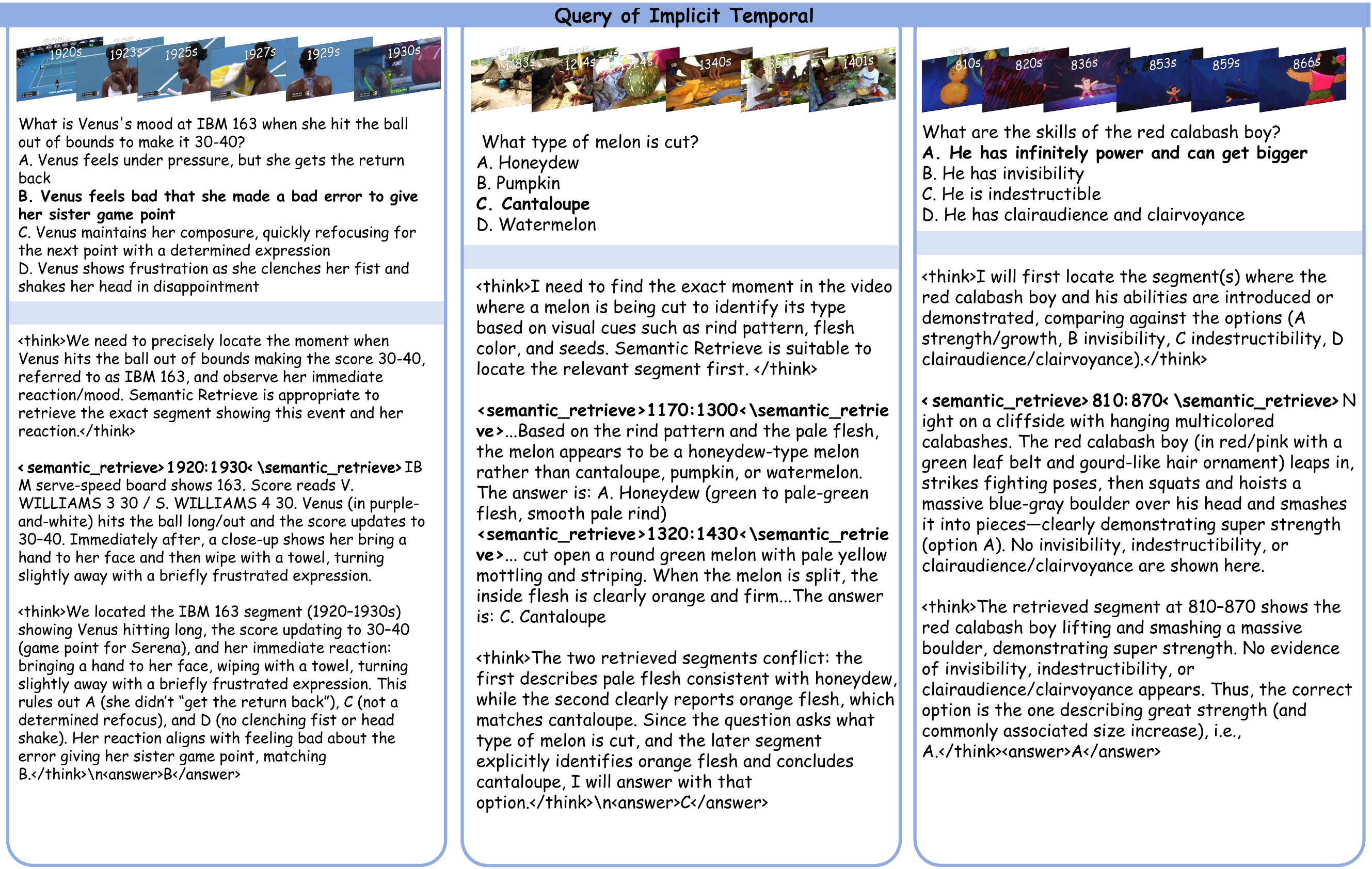}
    \caption{
    Additional case studies where implicit temporal queries favor Semantic Retrieve.
    When the answer depends on localized visual cues, such as a specific reaction, a brief object state, or the existence of a target event, Route2Look uses{Semantic Retrieve} to locate relevant segments before verification.
    }
    \label{fig:qualitative_implicit_retrieval}
\end{figure*}

\begin{figure*}[htbp]
    \centering
    \includegraphics[width=1.0\textwidth]{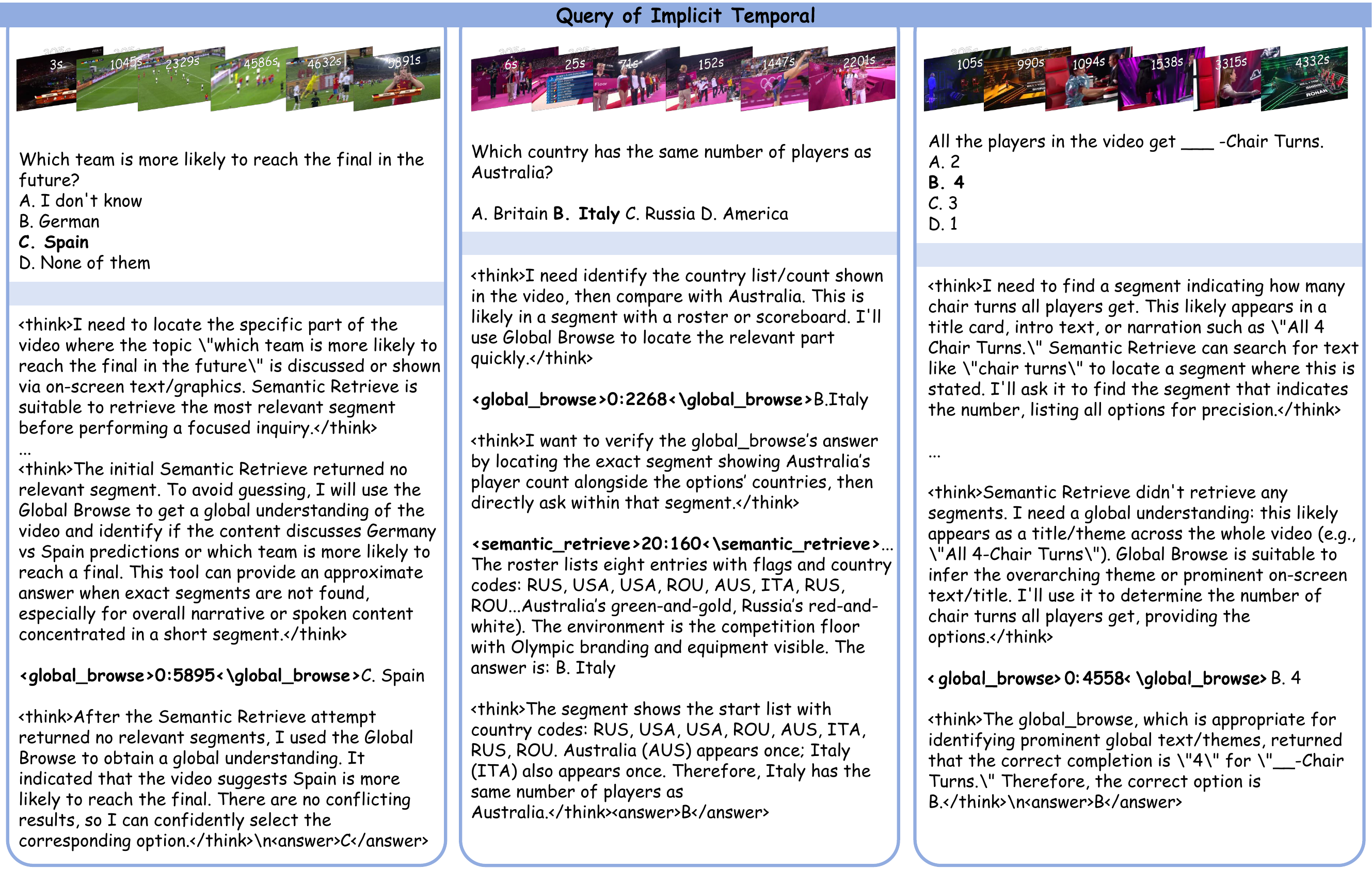}
    \caption{
    Additional case studies showing adaptive routing within implicit temporal queries.
    Route2Look can switch between {Semantic Retrieve} and {Global Browse} when the initial route is insufficient, showing adaptive routing within a single query.
    }
    \label{fig:qualitative_implicit_adaptive}
\end{figure*}

We provide additional qualitative examples in Fig.~\ref{fig:qualitative_global}--Fig.~\ref{fig:qualitative_implicit_adaptive} to further illustrate how Route2Look adapts evidence acquisition to different query types and evidence needs.
For global understanding queries in Fig.~\ref{fig:qualitative_global}, Route2Look directly invokes {Global Browse} to capture whole-video context, such as video category, documentary topic, and overall style.
For explicit temporal queries in Fig.~\ref{fig:qualitative_explicit}, it applies {Temporal Ground} to inspect the specified timestamp, time range, or temporal description, enabling direct verification of localized temporal evidence.

For implicit temporal queries, Route2Look exhibits more diverse routing behaviors.
When the query requires holistic judgment, repeated-event counting, global ordering, or long-range reasoning, as shown in Fig.~\ref{fig:qualitative_implicit_global}, Route2Look uses {Global Browse} to aggregate evidence across the full video.
When the answer depends on localized visual cues, such as a specific reaction, object state, or existence of a target event, it uses {Semantic Retrieve} to locate relevant temporal segments before verification, as shown in ~\ref{fig:qualitative_implicit_retrieval}.
In more ambiguous cases in ~\ref{fig:qualitative_implicit_adaptive}, Route2Look can switch between tools: it first attempts Semantic Retrieve for targeted evidence and then falls back to Global Browse or performs Semantic Retrieve when the initial Global Browse is insufficient.
These cases further show that implicit temporal queries do not favor a fixed strategy, and that Route2Look can adaptively route both across different queries and within a single query.

\begin{figure*}[htbp]
    \centering
    \includegraphics[width=1.0\textwidth]{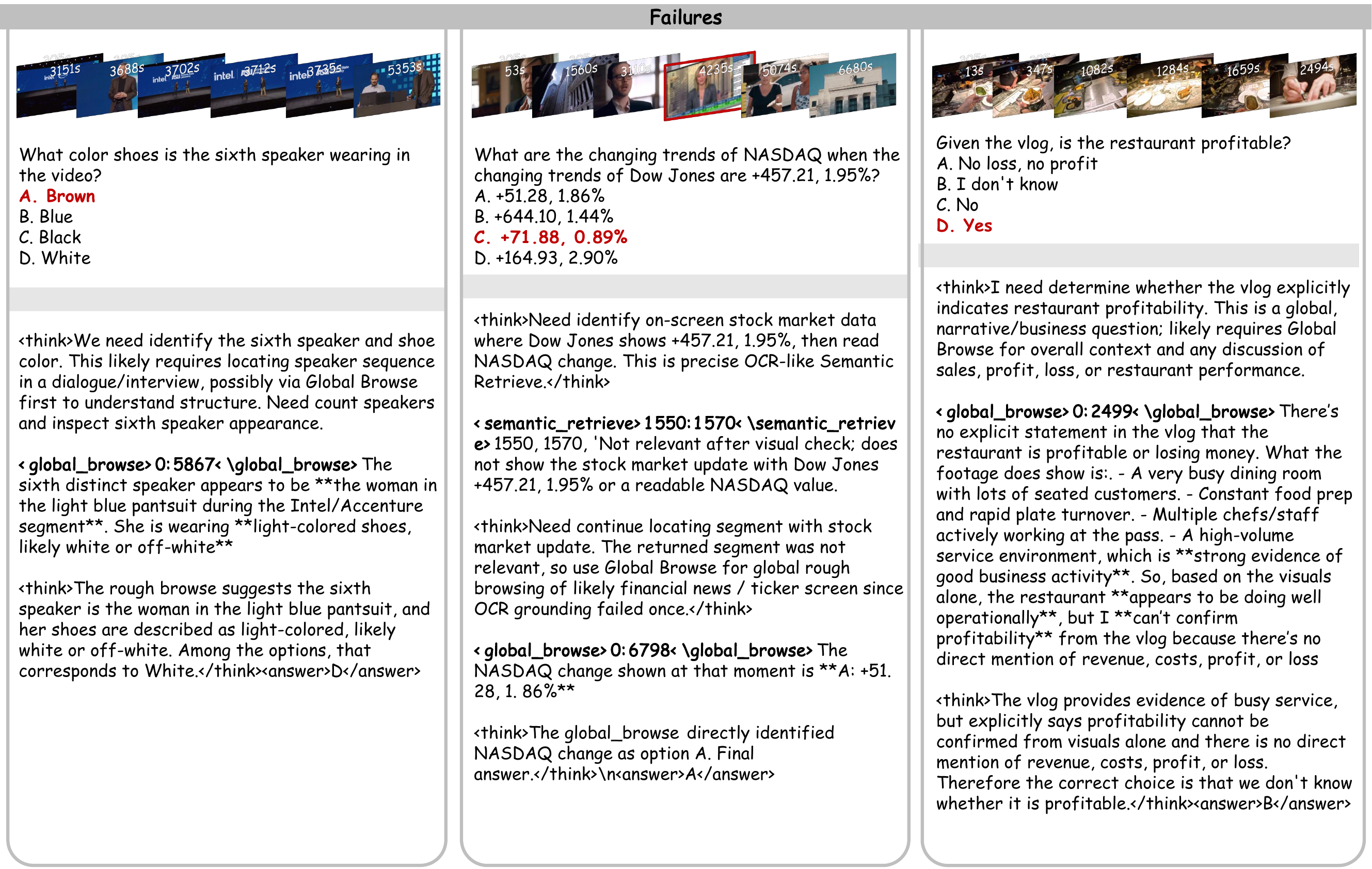}
    \caption{
    Representative failure cases.
    Route2Look may still fail on subtle spatial details, OCR-like numerical reading, or questions requiring reasoning beyond directly observable visual evidence.
    }
    \label{fig:qualitative_failures}
\end{figure*}

\end{document}